\documentclass[journal]{IEEEtran}
\usepackage{cite}
\usepackage{amsmath}

\usepackage{algorithm} 
\usepackage{algpseudocode}

\newcommand{\commentsymbol}{$\triangleright$}
\algrenewcommand\algorithmiccomment[1]{\hfill \commentsymbol{} \it #1} 

\ifCLASSOPTIONcompsoc
  \usepackage[caption=false,font=normalsize,labelfont=sf,textfont=sf]{subfig}
\else
  \usepackage[caption=false,font=footnotesize]{subfig}
\fi

\usepackage{graphicx}
\usepackage{gensymb}
\usepackage{amsfonts}
\usepackage{mathtools}
\usepackage{multirow}

\usepackage{amsthm}
\newtheorem{definition}{Definition}

\newcommand{\Bstrut}[1][1.2ex]{\rule[-#1]{0pt}{0pt}}
\newcommand{\Tstrut}[1][2.6ex]{\rule{0pt}{#1}}

\RequirePackage{dsfont}
\providecommand{\Z}{\ensuremath{\mathds{Z}} }
\providecommand{\R}{\ensuremath{\mathds{R}} }
\newcommand{\NNR}[0]{\R_{\geq 0}} 

\DeclareMathOperator*{\argmin}{arg\,min}

\DeclareFontFamily{OT1}{pzc}{}
\DeclareFontShape{OT1}{pzc}{m}{it}{<-> s * [1.10] pzcmi7t}{} %
\DeclareMathAlphabet{\mathpzc}{OT1}{pzc}{m}{it}
\newcommand{\fuzzy}[1]{\mathpzc{#1}}
\newcommand{\negphantom}[1]{\makebox[-\width]{\ensuremath{#1}}}
\newcommand{\vfuzzy}[1]{\text{$\fuzzy{#1}\negphantom{\fuzzy{#1}}\mkern0.6mu\fuzzy{#1}$}}
\def\dj{d\kern-0.4em\char"16\kern-0.05em}

\newcommand{\deuclidean}{\ensuremath{d_{E}}}

\newcommand{\card}[1]{\left|#1\right|}
\newcommand{\meanof}[1]{\text{avg}(#1)}

\newcommand{\avgerr}{\text{AE}}
\newcommand{\avgminerr}{\text{AME}}

\newcommand{\alphalevelsconstant}{\ell}

\newcommand{\fpoint}{\ensuremath{\fuzzy{p}}}
\newcommand{\fpointpos}{\ensuremath{p}}
\newcommand{\fset}{\ensuremath{\fuzzy{S}}}

\newcommand{\alphacut}[1]{\ensuremath{\prescript{\alpha\kern-0.1em}{}{#1}}}

\newcommand{\alphacutsub}[2]{\ensuremath{\prescript{\alpha_{#2}}{}{#1}}}

\newcommand{\fseta}{\ensuremath{\fuzzy{A}}}
\newcommand{\fsetb}{\ensuremath{\fuzzy{B}}}
\newcommand{\fsetp}{\ensuremath{\fuzzy{P}}}

\newcommand{\fsetv}{\ensuremath{\vfuzzy{S}}}
\newcommand{\fsetva}{\ensuremath{\vfuzzy{A}}}
\newcommand{\fsetvb}{\ensuremath{\vfuzzy{B}}}
\newcommand{\fpointv}{\ensuremath{\vfuzzy{\fpoint}}}

\newcommand{\set}[1]{\{#1\}\!}
\newcommand{\setcomplement}[1]{\ensuremath{\overline{#1}}}

\newcommand{\bidirmod}{\ensuremath{\dj}}

\newcommand{\damd}{\ensuremath{\protect\underrightarrow{d}_{\text{AMD}}}}
\newcommand{\damdr}{\ensuremath{\damd^{\scriptscriptstyle\!R}}}
\newcommand{\dsamd}{\ensuremath{d_{\text{AMD}}}}
\newcommand{\dsamdr}{\ensuremath{\dsamd^{\scriptscriptstyle R}}}

\newcommand{\dssamdr}{\ensuremath{\tilde{d}_{\text{AMD}}^{\scriptscriptstyle R}}}

\newcommand{\dalphacut}{\ensuremath{d^{\alpha}}}
\newcommand{\dalphacutbidir}{\ensuremath{\bidirmod^{\alpha}}}
\newcommand{\dalphacutsmd}{\ensuremath{d^{\alpha}_{\text{SMD}}}}
\newcommand{\dalphacutbidirsmd}{\ensuremath{\bidirmod^{\alpha}_{\text{SMD}}}}
\newcommand{\dalphacutwsmd}{\ensuremath{d^{\alpha}_{\text{wSMD}}}}
\newcommand{\dalphacutbidirwsmd}{\ensuremath{\bidirmod^{\alpha}_{\text{wSMD}}}}

\newcommand{\dalphabidirsamd}{\ensuremath{\bidirmod_{\alpha \text{AMD}}}}

\newcommand{\dalphabidirssamd}{\ensuremath{\tilde{\bidirmod}_{\alpha \text{AMD}}}}

\newcommand{\dalphaamdr}{\ensuremath{\protect\underrightarrow{d}{}^{\scriptscriptstyle\!\!R}_{\alpha \text{AMD}}}}
\newcommand{\dalphabidiramdr}{\ensuremath{\protect\underrightarrow{\bidirmod}{}^{\scriptscriptstyle\!\!R}_{\alpha \text{AMD}}}}
\newcommand{\dalphasamdr}{\ensuremath{d{}^{\scriptscriptstyle R}_{\alpha \text{AMD}}}}
\newcommand{\dalphabidirsamdr}{\ensuremath{\bidirmod{}^{\scriptscriptstyle R}_{\alpha \text{AMD}}}}
\newcommand{\dalphassamdr}{\ensuremath{\tilde{d}{}^{\scriptscriptstyle R}_{\alpha \text{AMD}}}}
\newcommand{\dalphabidirssamdr}{\ensuremath{\tilde{\bidirmod}{}_{\alpha \text{AMD}}^{\scriptscriptstyle R}}}

\newcommand{\setcardinality}[1]{\ensuremath{\left\vert{#1}\right\vert}}

\makeatletter
\newcommand*{\romucase}[1]{\expandafter\@slowromancap\romannumeral #1@}
\makeatother
\newcommand*{\romlcase}[1]{\romannumeral #1}

\makeatletter
\let\origsection\subsection
\renewcommand\subsection{\@ifstar{\starsection}{\nostarsection}}

\newcommand\nostarsection[1]
{\sectionprelude\origsection{#1}\sectionpostlude}

\newcommand\starsection[1]
{\sectionprelude\origsection*{#1}\sectionpostlude}

\newcommand\sectionprelude{%
	\vspace*{-0.4ex}
}

\newcommand\sectionpostlude{%
	\vspace*{-0.1ex}
}
\makeatother

\begin{document}

\title{Fast and Robust Symmetric Image Registration Based on Distances Combining Intensity and Spatial Information}

\author{Johan~\"{Of}verstedt,
        Joakim~Lindblad,~\IEEEmembership{Member,~IEEE,} 
        and~Nata\v{s}a~Sladoje,~\IEEEmembership{Member,~IEEE}
\thanks{The authors are with the Centre for Image Analysis, Department of Information Technology, Uppsala University, Uppsala, Sweden. Lindblad and Sladoje are also with Mathematical Institute of the Serbian Academy of Sciences and Arts, Belgrade, Serbia.  E-mail: \{johan.ofverstedt,joakim.lindblad,natasa.sladoje\}@it.uu.se}
}

\markboth{}
{\"{Of}verstedt \MakeLowercase{\textit{et al.}}: Fast Image Registration Using a Symmetric Distance Measure Based on Intensity and Spatial Information}

\maketitle

\begin{abstract}
Intensity-based image registration approaches rely on similarity measures to guide the search for geometric correspondences with high affinity between images. The properties of the used measure are vital for the robustness and accuracy of the registration. In this study a symmetric, intensity interpolation-free, affine registration framework based on a combination of intensity and spatial information is proposed. The excellent performance of the framework is demonstrated on a combination of synthetic tests, recovering known transformations in the presence of noise, and real applications in biomedical and medical image registration, for both 2D and 3D images. The method exhibits greater robustness and higher accuracy than similarity measures in common use, when inserted into a standard gradient-based registration framework available as part of the open source Insight Segmentation and Registration Toolkit (ITK). The method is also empirically shown to have a low computational cost, making it practical for real applications. Source code is available.
\end{abstract}


\IEEEpeerreviewmaketitle

\section{Introduction}

\IEEEPARstart{I}{MAGE} registration \cite{MAINTZ19981,VIERGEVER2016140,zitova2003image,oliveira2014medical} is the process of establishing correspondences between images and a reference space, such that the contents of the images have a high degree of affinity in the reference space. An example of such correspondence is a mapping of an image (often referred to as {\it floating} image) of a brain to a reference space of another image (often referred to as {\it reference} image) of a brain where their important structures are well co-localized. There are two main categories of approaches for image registration: {\it feature-based} methods extract a set of feature points between which a correspondence is found, whereas {\it intensity-based} methods use the voxel-values directly, and evaluate candidate mappings based on a similarity measure (affinity). There are also two main categories of transformation models: linear (which include, as special cases, rigid, similarity, and affine transformations), and non-linear (deformable). The combination of differentiable transformation models and differentiable similarity measures enables the use of gradient-based local optimization methods.

Medical and biomedical image registration, \cite{sotiras2013deformable,oliveira2014medical,matl2017vascular}, is an important branch of general image registration and a lot of effort has been invested over the last decades to refine the tools and techniques, \cite{VIERGEVER2016140}. Although a majority of the recent research has been devoted to non-linear registration techniques, the most prevalent registration method used in the clinic is still linear registration. In a number of situations, the deformations allowed by non-linear registration can be difficult to evaluate and may affect reliability of diagnosis, \cite{VIERGEVER2016140}; hence, physicians may prefer a more constrained rigid or affine alignment. Considering their wide usage as fundamental tools, improvement of rigid and affine registration in terms of performance and reliability is highly relevant in practice.

\begin{figure*}[t!]
\centering
\subfloat[Reference Image\label{refcilia}]{\hspace{0.4cm}\includegraphics[width=2.5cm, height=2.5cm]{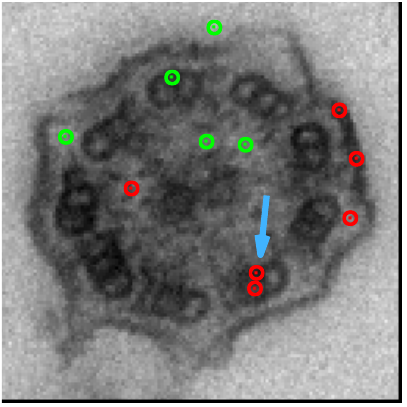}\hspace{0.4cm}}
\subfloat[Floating Image\label{flocilia}]{\hspace{0.4cm}\includegraphics[width=2.5cm, height=2.5cm]{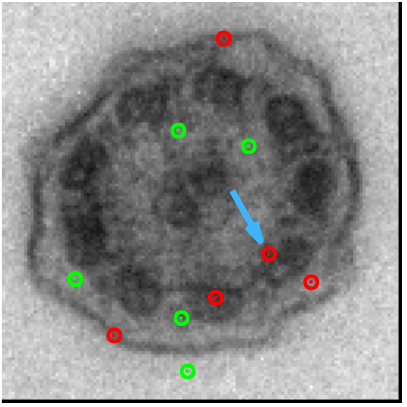}\hspace{0.4cm}}
\subfloat[FB Rigid (Fail)\label{rigidcilia}]{\hspace{0.4cm}\includegraphics[width=2.5cm, height=2.5cm]{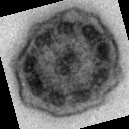}\hspace{0.4cm}}
\subfloat[FB Affine (Fail) \label{affinecilia}]{\hspace{0.4cm}\includegraphics[width=2.5cm, height=2.5cm]{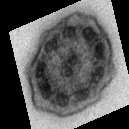}\hspace{0.4cm}}
\subfloat[IB, Rigid+Affine\label{intensitybasedcilia}]{\hspace{0.4cm}\includegraphics[width=2.5cm, height=2.5cm]{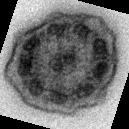}\hspace{0.4cm}}
\caption{Illustrative example of a biomedical registration task where a widely used feature-based (FB) method (SIFT, as implemented in FIJI-plugin Linear Stack Alignment\protect\footnotemark) fails, while the proposed intensity-based (IB) method (Sec. \ref{sect:ciliaregistration}) performs well. Green points in (a) and (b) are incorrectly detected as having a match, and red points do not have a match. The feature-extractor fails to detect points corresponding to the relevant structures (one approximately correct match, indicated with arrows, can be found manually), and both the central rings and the outer rings are misaligned.}
\label{fig:featurefail}
\end{figure*}

Feature-based image registration is dependent on the ability of the feature extraction method to locate distinct points of interest appearing in both (all) images. Feature-extractors (e.g. SIFT \cite{lowe1999object}) typically presuppose the existence and relevance of specific local characteristics such as edges, corners and other salient features; if no, or too few, such distinct points are found, the registration will fail. This is often the case in medical and biomedical applications, \cite{BERKELS201446,fischer2008ill}, where intensity-based registration, therefore, tends to be the method of choice. Figure~\ref{fig:featurefail} shows an illustrative example of a biomedical application where a feature-based method fails, whereas an intensity-based method can be successful.

Intensity-based registration is, in general, formulated as a non-convex optimization problem. The  similarity measures commonly used as optimization criteria  typically exhibit a high number of local optima \cite{skerl2006protocol,lindblad2014linear}; a count which tends to rapidly increase under noisy conditions. A small region of attraction of a global optimum imposes that the starting position has to be set very close to the optimal solution for it to be found by an optimizer. This leads to reliability challenges for automated solutions.

In this study we develop a registration framework based on a family of symmetric distance measures, proposed in  \cite{lindblad2014linear}, which combine intensity and spatial information in a single measure. These measures have been shown to be characterized by smooth distance surfaces with significantly fewer local minima than the commonly used intensity-based measures, when studied in the context of template matching and object recognition.
In this work we demonstrate that slightly modified versions of these distance measures can be successfully used for fast and robust affine image registration. By differentiating the distance measure we are able to use efficient gradient-based optimization.
The proposed method outperforms the commonly used similarity measures in both synthetic and real scenarios of medical and biomedical registration tasks, which we confirm by (i) landmark-based evaluation on transmission electron microscopy (TEM) images of cilia \cite{suveer2017enhancement}, with the aim of improving multi-image super-resolution reconstruction, as well as (ii) evaluation on the task of atlas-based segmentation of magnetic resonance (MR) images of brain, on the LPBA40-dataset \cite{shattuck2008construction}.

Intensity interpolation is typically a required tool in the context of intensity-based registration performed with commonly used similarity measures since the sought transformation (and intermediate candidates) is likely to map points to regions outside of the regular grid. Treating the reference and floating images differently in terms of the interpolation introduces a significant source of asymmetry \cite{avants2014insight} and may lead to success or failure of a registration task depending on which image is selected as \emph{reference} and which is \emph{floating}. 
Our proposed approach requires no off-grid intensity values, and is interpolation-free in terms of intensities; empirical tests confirm that it is highly symmetric in practice.

Noting that intensity-based image registration can be computationally demanding, we also include a study of execution time of (i) isolated distance and gradient computations through micro-benchmarks, and (ii) entire image registration tasks. We observe that the proposed measure is fast to compute in comparison with the implementations of the measures existing in the ITK-framework \cite{avants2014insight}. The proposed registration framework is implemented in C++/ITK, as well as in Python/NumPy/SciPy, and its source code is available\footnotetext{imagej.net/Linear\_Stack\_Alignment\_with\_SIFT}\footnote{Source code available from www.github.com/MIDA-group}.

\section{Preliminaries and Previous Work}

\subsection{Images as Fuzzy Sets}

First we recall a few basic concepts related to fuzzy sets \cite{zadeh1965information}, a theoretical framework where gray-scale images are conveniently represented.

A \emph{fuzzy set} $\fset$ on a reference set $X_{\fset}$ is a set of ordered pairs,
$\displaystyle{\fset = \set{(x,\mu_{\fset}(x)) \, \colon x \in X_{\fset} }\,}$,
where $\mu_{\fset} \colon X_{\fset} \to [0, 1]$ is the \emph{membership function} of $\fset$.
Where there is no risk for confusion, we equate the set and its membership function and let $\fset(x)$ be equivalent to $\mu_\fset(x)$.

A gray-scale image can directly be interpreted as a spatial fuzzy set by rescaling the valid intensity range to $[0,1]$. We assume,  w.l.o.g., that the images to be registered have an intensity range $[0,1]$ and we directly interpret them as fuzzy sets defined on a reference set which is the image domain, and is in most cases a subset of $\Z^n$. We use the terms \emph{image} and \emph{fuzzy set} interchangeably in this text.

A crisp set $C \subseteq X_C$ (a binary image) is a special case of a fuzzy set, with its characteristic function as membership function
\begin{equation}
\mu_C(x) = 
\left\{\begin{array}{ll}
        1, \;\; & \text{for } \,\, x \in C\\
        0, \;\; & \text{for } \,\, x \notin C\,.
        \end{array}\right.
\end{equation}
The \emph{height} of a fuzzy set $S \subseteq X_{\fset}$ is $\;\displaystyle h(\fset) = \max_{x \in X_{\fset}} \mu_S(x)$.
The \emph{complement} $\setcomplement{\fset}$ of a fuzzy set $\fset$ is $\setcomplement{\fset} = \set{(x, 1-\mu_{\fset}(x)) \, \colon x \in X_{\fset}} \, .$ 
An \emph{$\alpha$-cut} of a fuzzy set $\fset$ is a crisp set defined as $\;\alphacut{\fset}~=~\set{x \in X_{\fset} : \mu_{\fset}(x) \geq \alpha}\,,$ i.e., a thresholded image.

Let $p$ be an element of the reference set $X_{\fset}$. A \emph{fuzzy point} $\fpoint$ (also called a fuzzy \emph{singleton}) defined at $p \in X_{\fset}$ with height $h(\fpoint)$, is defined by a membership function
\begin{equation}
\mu_{\fpoint}(x) = 
\left\{\begin{array}{ll}
        h(\fpoint), \;\; & \text{for } \,\, x = \fpointpos\\
        0, \;\; & \text{for } \,\, x \neq \fpointpos\,.
        \end{array}\right.
\end{equation}

\subsection{Intensity-Based Registration and Point-Wise Distances}\label{sec:pointWiseDist}

Intensity-based registration is a general approach to image registration defined as a minimization process, where a distance measure between the intensities of overlapping points (or regions) is used  as optimization criterion. Given a distance measure $d$ and a set of valid transformations $\Omega$, intensity-based registration of two images $\fseta$ (floating) and $\fsetb$ (reference) can be formulated as the optimization problem, 
\begin{equation}
\hat{T} = \argmin_{T \in \Omega} d(T(\fseta), \fsetb),
\label{eqregmain}
\end{equation}
where $T(\fseta)$ denotes a valid transform of image $\fseta$ into the reference space of image $\fsetb$. 

Intensity-based similarity/distance measures which are most commonly used for image registration are Sum of Squared Differences (SSD) \cite{HAJNAL1995Registration}, 
Pearson Correlation Coefficient (PCC)
and Mutual Information (MI) \cite{viola1997alignment}. These measures are point-based, i.e. they are  functions of the intensities of points belonging to the overlapping regions of the two compared sets. Their evaluation, therefore, typically 
requires interpolation of image intensities.

For two images $P$ and $Q$ defined on a common reference set $X_{P,Q}$ of overlapping points,
these measures are defined as 
\begin{equation}
\textrm{SSD}(P, Q) = \sum_{v \in X_{P, Q}}(P(v)-Q(v))^2,
\end{equation}
\begin{equation}
\label{PCC}
\textrm{PCC}(P,Q)\!=\!\frac{\sum\limits_{v \in X_{P,Q}}\!\!(P(v)-{\meanof{P}})(Q(v)-{\meanof{Q}})}{ \sqrt{\sum\limits_{v \in X_{P,Q}\hspace*{-1em}}\!\!(P(v)\!-\!{\meanof{P}})^2} \!\sqrt{\sum\limits_{v \in X_{P,Q}\hspace*{-1em}}\!\!(Q(v)\!-\!{\meanof{Q}})^2} }
\end{equation}
and 
\begin{equation}
\label{MI}
\text{MI}(P, Q) = H_P + H_{Q} - H_{P, Q}.
\end{equation}
In~\eqref{PCC} ${\meanof{P}}$, and ${\meanof{Q}}$ denote means of the resp. intensity distributions over the evaluated region. 
In~\eqref{MI} the (joint and marginal) entropies $H_P$, $H_Q$ and $H_{P,Q}$ of the image intensity distributions $P$ and $Q$
are defined in terms of the estimated probability $p$ of a randomly selected point $v$ having intensities $P(v)$, $Q(v)$, as
\begin{equation}
H_P = -\!\!\sum_{v\in X_{P,Q}}\!\! p(P(v))\log(p(P(v)))\,,
\end{equation}
and
\begin{equation}
H_{P, Q} = -\!\!\sum_{v\in X_{P,Q}}\!\! p(P(v), Q(v))\log(p(P(v), Q(v)))\,.
\end{equation}

Intensity-based registration, as formulated in \eqref{eqregmain}, is, in general, a non-convex optimization problem with a large number of local optima, especially for the commonly used point-based measures (SSD, PCC, and MI). To try to overcome this optimization challenge, a resolution-pyramid-scheme is normally used \cite{irani1991improving,rosenfeld2013multiresolution}, where smoothed low resolution images are first registered, followed by registration of images with increasing resolution and decreasing degree of smoothing, using the transform obtained from the previous stage as starting position (so-called coarse-to-fine approach).

\subsection{Distances Combining Intensity and Spatial Information}

While the distances of Sec. \ref{sec:pointWiseDist} only rely on intensities of overlapping points, the distances presented in this section incorporate also spatial information of non-overlapping points. 
For such spatial relations, we consider distances between two points, between a point and a set, and between two sets. The most commonly used point-to-point distance is the Euclidean distance, denoted $\deuclidean$. 

Given a point-to-point distance $d(a, b)$, the common crisp point-to-set distance between a point $a$ and a set $B$ is
\begin{equation}
\label{eq:crisp_pts}
  d(a, B) = \inf\limits_{b \in B}{d(a, b)}\,.
\end{equation}
Closely related to the crisp point-to-set distance is the (external) distance transform of a crisp set $B \subseteq X_B$ (with point-to-point distance $d$) which is defined as
\begin{equation}
\text{DT}[B](x) = \min_{y \in B} \set{d(x, y)} \;.
\end{equation}

Taking into the consideration the intensity, or equivalently, the \emph{height} of a fuzzy point, 
the \emph{fuzzy point-to-set inwards distance} $d^\alpha$, based on integration over $\alpha$-cuts \cite{lindblad2014linear}, between a fuzzy point $\fpoint$ and a fuzzy set $\fset$, is defined as
\begin{equation}
\label{eq:dalpha}
d^{\alpha}(\fpoint, \fset) = \int_0^{h(\fpoint)} d(\fpointpos, \alphacut{\fset}) \, \mathrm{d}\alpha\,,
\end{equation}
where $d$ is a point-to-set distance defined on crisp sets. The \emph{complement distance} \cite{lindblad2009set} of a fuzzy point-to-set distance $d$ is
\begin{equation}
\setcomplement{d}(\fpoint, \fset) = d(\setcomplement{\fpoint}, \setcomplement{\fset})\,.
\end{equation}
The \emph{fuzzy point-to-set bidirectional distance} $\bidirmod^\alpha$ is
\begin{equation}
\label{eq:djalpha}
\bidirmod^\alpha(\fpoint, \fset) = d^\alpha(\fpoint, \fset) + \setcomplement{d}{^\alpha}(\fpoint, \fset)\,.
\end{equation}

For an arbitrary point-to-set distance $d$, Sum of Minimal Distances (SMD) \cite{eiter1997distance} defines a set-to-set distance as 
\begin{equation}
\label{eq:smd}
d_{\text{SMD}}(A, B) = \frac{1}{2}\Big(\sum\limits_{a \in A} d(a, B) + \sum\limits_{b \in B} d(b, A)\Big)\,.
\end{equation}
A weighted version can be defined \cite{lindblad2014linear}, which may be useful if a priori information about relative importance of contributions of different points to the overall distance is available:
\begin{equation}
\label{eq:wsmd}
\begin{split}
d_{\text{wSMD}}(A, B; w_A, w_B) = \\
\frac{1}{2}\Big(\sum\limits_{a \in A} w_A(a) d(a, B) + \sum\limits_{b \in B} w_B(b) d(b, A)\Big)\,.
\end{split}
\end{equation}

Inserting distances \eqref{eq:dalpha} or \eqref{eq:djalpha} in \eqref{eq:smd} or \eqref{eq:wsmd}
provides extensions of the SMD family of distances to fuzzy sets~\cite{lindblad2014linear}. We refer to them as $\dalphacutsmd$, $\dalphacutbidirsmd$, $\dalphacutwsmd$ and $\dalphacutbidirwsmd$.

It has been observed  for fuzzy set distances \cite{brass2002nonexistence} in general, and for distances based on \eqref{eq:dalpha} and \eqref{eq:djalpha} in particular, that distances between sets with empty $\alpha$-cuts may give infinite or ill-defined distances. We follow a previous study and introduce a parameter $d_{\text{MAX}} \in \R_{\ge 0}$, \cite{ofverstedt2017distance}, to limit the underlying crisp point-to-set distance. This has a double benefit of (i) reducing the effect of outliers and (ii) making the distances well defined also for images with empty $\alpha$-cuts.

Distances based on Optimal Mass Transport (OMT), such as the Wasserstein distance, also combine intensity and spatial information, and are widely studied and used in image processing \cite{rubner2000earth}.
The OMT can be framed as a linear programming optimization problem, which is solvable in $\mathcal{O}(N^3)$ \cite{wang2013linear}. This is intractable for most realistic image processing scenarios, and approximations are typically considered \cite{wang2013linear},\cite{pele2009fast}. It is possible to incorporate these distances in image registration frameworks, but to the best of our knowledge, this has only been done for non-linear (deformable) registration, and has been shown to be very computationally demanding \cite{haker2004optimal,ur20093d}. We performed a preliminary study of OMT-based methods using the formulation in \cite{pele2009fast}, and observed both very high computational demands and noisy distance landscapes. In absence of a complete registration framework for linear registration based on OMT, this family of measures is excluded from the empirical part of this study.

\subsection{Transformations, Interpolation, and Symmetry}

Linear transformations relate points in one space to another via application of a linear function. 
A transformation is rigid if only rotations and translations are permitted, and affine if shearing and reflections are also permitted. Affine transformation $T:\R^n \to \R^n$ can be expressed as
matrix multiplication,
\begin{equation}
\label{eq:transf}
Tx = \scalebox{0.8}{$ 
\begin{bmatrix}
    a_{11} & a_{12} & \dots & a_{1n} & t_{1} \\
    a_{21} & a_{22} & \dots & a_{2n} & t_{2} \\
    \vdots & \vdots & \ddots & \vdots & \vdots \\
    a_{n1} & a_{n2} & \dots & a_{nn} & t_{n} \\
    0       & 0 & \dots & 0 & 1
\end{bmatrix}
\begin{bmatrix}
 x_1 \\
 x_2 \\
 \vdots \\
 x_n \\
 1
\end{bmatrix}$}.
\end{equation}

Linear transformations can, in general, transform points sampled on an image grid to positions outside of the grid, hence an interpolator is required for obtaining the image intensity at the transformed point's location. Interpolation is a large source of error, bias, and a significant contributing factor of asymmetry in intensity-based registration, \cite{avants2014insight}. Commonly, interpolation is only required for one of the two images, where sampling (for optimization) is done from the grid of the other image space; hence, the two images are treated asymmetrically, yielding distinct similarity surfaces (over the transformation parameters) depending on which image is taken as reference. This can cause a registration task to succeed or fail, depending on the registration direction.

\subsection{Optimization}

Registration with a differentiable distance measure as objective function enables the use of gradient-based optimization algorithms, which typically are significantly more efficient than derivative-free algorithms for local iterative optimization. An effective and commonly used subset of gradient-based algorithms are the stochastic gradient descent methods \cite{ruder2016overview}, which consider a random subset of the points in each optimization iteration, incurring a two-fold benefit: utilizing randomness to escape shallow local optima in the implicit distance surface, while also decreasing the computational work required per iteration. The size of the random subset is usually given as a fraction of the total number of points, and denoted as the sampling fraction. Approximation of the cost function by random subset sampling (where a new random subset of points is chosen in every iteration) has been,  in previous studies, \cite{viola1997alignment,klein2007evaluation}, shown to perform well for intensity-based registration.

\section{Proposed Image Registration Framework}

\subsection{Distances}

To extend the family of distance measures  given by \eqref{eq:wsmd}, to be suitable for registration, optionally with random subsampling optimization methods \cite{klein2007evaluation}, we here define a new related family of distance measures.

\begin{definition}[Asymmetric average minimal distance]
\label{def:dalphaamd_plain}
Given fuzzy set $\fseta$ on a reference set $X_\fseta\!\subset\!\R^n$, fuzzy set $\fsetb$ on reference set $X_\fsetb\!\subset\!\R^n$, and a weight function $w_A\colon X_\fseta\!\to\!\NNR$, 
the \emph{Asymmetric average minimal distance} from $\fseta$ to $\fsetb$, 
is
\begin{equation}
\label{eq:dalphaamd_plain}
\damd(\fseta, \fsetb; w_A) = 
\frac{1}{\sum\limits_{x \in X_{\fseta}}\!w_A(x)} \sum\limits_{x \in X_{\fseta}}\!{w_A(x)d(\fseta(x), \fsetb)}\,.
\end{equation}%
\end{definition}
\noindent
We consider point-to-set distances defined by \eqref{eq:dalpha} or \eqref{eq:djalpha}. 

Building on the asymmetric distance, 
we formulate a symmetric distance as follows:

\begin{definition}[Average minimal distance]
\label{def:dalphasamd_plain}
Given fuzzy set $\fseta$ on reference set $X_\fseta \subset \R^n$, fuzzy set $\fsetb$ on reference set $X_\fsetb \subset \R^n$, weight functions $w_A\colon X_\fseta \to \NNR$ and $w_B\colon X_\fsetb \to \NNR$, 
the \emph{Average minimal distance} between $\fseta$ and $\fsetb$,  
is
\begin{equation}
\label{eq:dalphasamd_plain}
\begin{split}
\dsamd(\fseta, \fsetb; w_A, w_B) = \\
\frac{1}{2}\big( \damd(\fseta, \fsetb; w_A) +
\damd(\fsetb, \fseta; w_B) \big)\,.
\end{split}
\end{equation}
\end{definition}

In the context of image registration, we utilize $\damd$ to express a (weighted) distance between transformed fuzzy points $T(\fseta(x))$, and the image $\fsetb$, where the transformation of a fuzzy point $\fseta(x)=\set{(x,\mu_\fseta(x))}$ is given by the transformation of the reference point $x$:
\begin{equation}
T(\fseta(x)) = \set{(T(x), \mu_\fseta(x))}\;.
\end{equation}

\begin{figure}[t!]
    \centering
    \subfloat[$\fseta$ with weights $w_A$ (radius).]{\includegraphics[width=4cm,trim={5mm 6mm 3mm 9mm},clip]{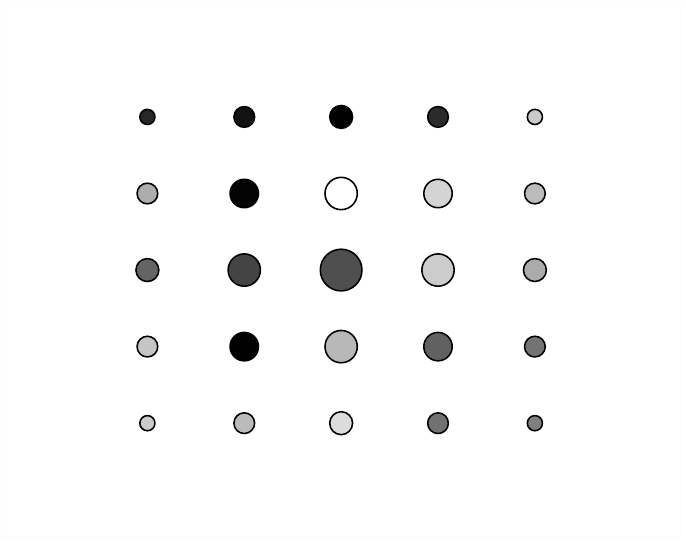}}\hspace{0.5cm}
    \subfloat[$\fsetb$ with mask $M_B$ (surrounding rectangle).]{\includegraphics[width=4cm,trim={5mm 6mm 3mm 9mm},clip]{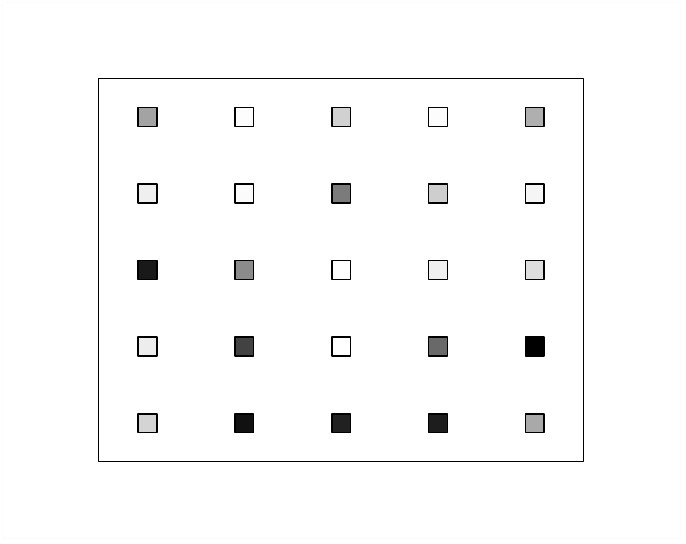}}\hspace{0.5cm}\\
    \subfloat[$\fseta$ mapped into the space of $\fsetb$. Triangles mark points mapped outside $M_B$, and thus discarded.]{\includegraphics[width=4cm,trim={5mm 5mm 3mm 6mm},clip]{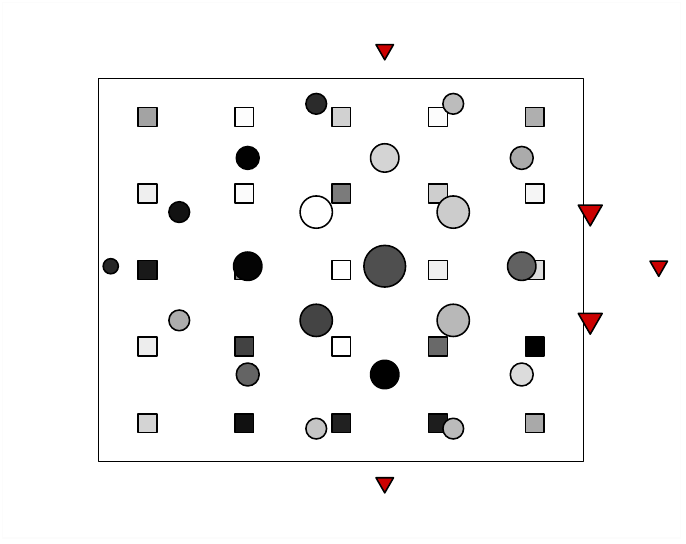}}\hspace{0.5cm}
    \subfloat[Distance contribution graph for $\dalphacutbidir(T(\fseta(x)), \fsetb)$; where $x$ is central point of $\fseta$.]{\includegraphics[width=4cm,trim={5mm 5mm 3mm 6mm},clip]{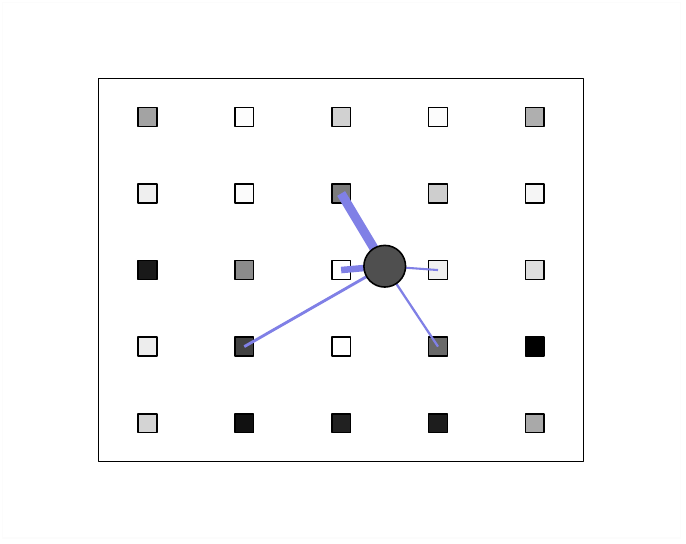}}\\
    \subfloat[Inwards part of $\dalphacutbidir$: $\dalphacut(\fseta(x), \fsetb)$ point-to-set distance.]{\includegraphics[width=8.0cm,height=2cm]{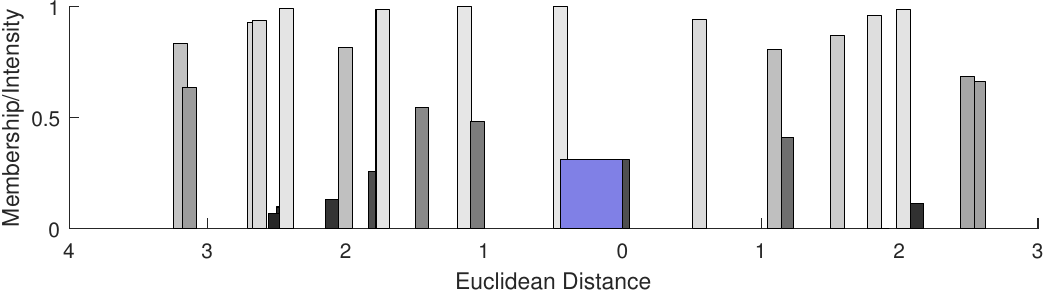}}\hspace{0.5cm}\\
    \subfloat[Complement part of $\dalphacutbidir$: $\dalphacut(\setcomplement{\fseta(x)}, \setcomplement{\fsetb})$ point-to-set distance.]{\includegraphics[width=8.0cm,height=2cm]{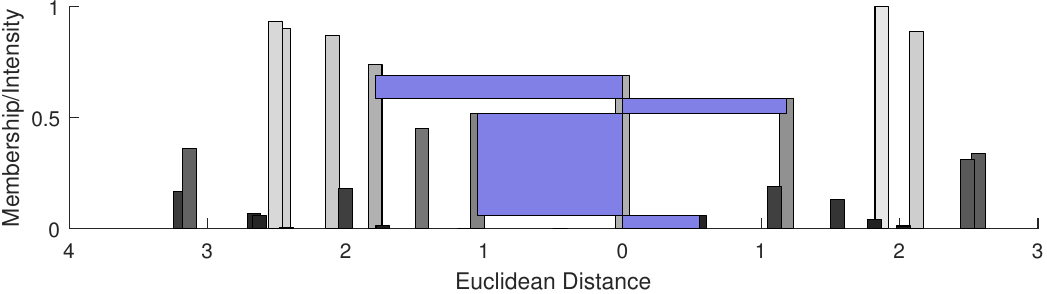}}
    \caption{Illustration of the Asymmetric average minimal distance for image registration. (a) Source set $\fseta$ (radius represents associated weight; gray-level represents membership). (b) Target set $\fsetb$ with associated mask $M_B$. (c) The transformed (by rotation and translation) $\fseta$ on top of $\fsetb$. (d) Illustration of the contributions to the point-to-set distance $\dalphacutbidir$ by the central point of $\fseta$. Thickness of lines show the $\alpha$-integrated height contributed by each point in $\fsetb$. (e-f) The inwards and complement parts of $\dalphacutbidir$ visualized as 1D graphs, where the x-axis is the Euclidean distance (in 2D) from the mid-point and the points at the left and right side (of the origin) respectively are the points on the left and right side of the mid-point $T(\fseta(x))$ in (d).}
    \label{fig:damd_demo}
\end{figure}

To reflect the bounded image domain, only the transformed points falling on a predefined region $M_B\subset\R^n$
are considered. Note that, when $\fseta$ and $\fsetb$ are digital images, $X_\fseta$ and $X_\fsetb$ are typically subsets of $\Z^n$ and the transformed points $T(x)\rvert_{x\in X_\fseta}$ do not necessarily coincide with the points of the reference set $X_\fsetb$; an illustrative example is given in Fig.~\ref{fig:damd_demo}. 

We, therefore, provide the following definitions suited for the task of image registration: 
\begin{definition}[Asymmetric average minimal distance for image registration]
\label{def:dalphaamdr}
Given fuzzy set $\fseta$ on reference set $X_\fseta \subset \R^n$, fuzzy set $\fsetb$ on $X_\fsetb \subset \R^n$, a weight function $w_A\colon X_\fseta \to \NNR$, and a crisp subset (mask) $M_B \subset \R^n$, the \emph{Asymmetric average minimal distance for image registration} from $\fseta$ to $\fsetb$, parameterized by a transformation $T\colon X_\fseta \to \R^n$, is
\begin{equation}
\label{eq:dalphaamdr}
\begin{split}
\damdr(\fseta, \fsetb; T, w_A, M_B) = \\
\frac{1}{\sum\limits_{x \in \hat{X}}\!{w_A(x)}} \sum\limits_{x \in \hat{X}}^{}{w_A(x)d(T(\fseta(x)), \fsetb)}\\
\textrm{where } \hat{X} = \set{x \, \colon x \in X_{\fseta} \wedge T(x) \in M_B}\;. 
\end{split}
\end{equation}
\end{definition}

\begin{definition}[Average minimal distance for image registration]
\label{def:dalphasamdr}
Given fuzzy set $\fseta$ on reference set $X_\fseta$, fuzzy set $\fsetb$ on $X_\fsetb$, weight functions $w_A\colon X_\fseta \to \NNR$ and $w_B\colon X_\fsetb \to \NNR$, and crisp subsets (masks) $M_A,M_B \subset \R^n$, the \emph{Average minimal distance for image registration} between $\fseta$ and $\fsetb$, parameterized by an invertible transformation $T\colon \R^n \to \R^n$, with inverse $T^{-1}$, is defined as
\begin{equation}
\label{eq:dalphasamdr}
\begin{split}
\dsamdr(\fseta, \fsetb; T, w_A, w_B, M_A, M_B) = \\
\frac{1}{2}\big(\!\damdr(\fseta, \fsetb; T, w_A, M_B) + \!\damdr(\fsetb, \fseta; T^{-1}\!, w_B, M_A) \big)\,.
\end{split}
\end{equation}
\end{definition}

The distance $\dsamdr$ is based on full sampling, taking into account all points in the two sets which have non-zero weights,  as long as they are transformed to points inside the mask associated with the other set. To reduce the computational cost of the distance and, in addition, to enable random iterative sampling, we propose an approximate version of  $\dsamdr$:  

\begin{definition}[Subsampled average minimal distance for image registration]
\label{def:dalphassamdr}
Given fuzzy set $\fseta$ on reference set $X_\fseta$, fuzzy set $\fsetb$ on $X_\fsetb$, weight functions $w_A\colon X_\fseta \to \NNR$ and $w_B\colon \fsetb \to \NNR$, and crisp subsets (masks) $M_A, M_B \subset \R^n$, the \emph{Subsampled average minimal distance for image registration} between $\fseta$ and $\fsetb$,  parameterized by an invertible transformation $T\colon \R^n \to \R^n$, with inverse $T^{-1}$, and crisp sets $P_A \subseteq X_\fseta$ and $P_B \subseteq X_\fsetb$, is defined as
\begin{equation}
\label{eq:dalphassamdr}
\begin{split}
\dssamdr(\fseta, \fsetb; \fsetp_A, \fsetp_B, T, w_A, w_B, M_A, M_B) = \\
\frac{1}{2}\big( \damdr(\fseta\cap P_A, \fsetb; T, w_A, M_B) +\\
\damdr(\fsetb\cap P_B, \fseta; T^{-1}, w_B, M_A) \big)\,.
\end{split}
\end{equation}
\end{definition}

Inserting \eqref{eq:dalpha} or \eqref{eq:djalpha} as point-to-set distance in \eqref{eq:dalphaamdr}, and hence indirectly in \eqref{eq:dalphasamdr} and \eqref{eq:dalphassamdr}, provides extensions of this family of distances to the $\alpha$-cut-based distances, which we denote $\dalphaamdr$, $\dalphabidiramdr$, $\dalphasamdr$, $\dalphabidirsamdr$, $\dalphassamdr$ and $\dalphabidirssamdr$.

Normalization of the weights of the sampled points, introduced through Def. \ref{def:dalphaamd_plain}, renders the magnitude of the distance (and subsequently its derivatives) invariant to the size and aggregated weight of the sets or of the sampled subsets. Since the normalization is done separately from $\fseta$ to $\fsetb$ and from $\fsetb$ to $\fseta$, both directions are weighted equally even if the total weights of the point subsets from the two sets are different. This normalization can simplify the process of choosing e.g. step-length of various optimization methods, and makes it more likely that default hyper-parameter values can be found and reused across different applications.

\subsection{Registration}

We propose to utilize symmetric distances $\dalphabidirsamd$ and $\dalphabidirssamd$ as cost functions in \eqref{eqregmain} to define  concrete image registration methods. Inserting $\dalphabidirsamdr$ into \eqref{eqregmain}  we obtain
\begin{equation}
\label{eqregmain2}
\hat{T} = \argmin_{T \in \Omega} \dalphabidirsamdr(\fseta, \fsetb; T, w_A, w_B, M_A, M_B)\,.
\end{equation}

For the case of subset sampling with sets $P_A$ and $P_B$, registration is defined as
\begin{equation}
\label{eqregmain3}
\hat{T} =
\argmin_{T \in \Omega} \dalphabidirssamdr(\fseta, \fsetb; P_A, P_B, T, w_A, w_B, M_A, M_B).
\end{equation}
By selection of new random subsets $P_A$ and $P_B$ in each iteration, various stochastic gradient descent optimization methods can be realized.

To solve the optimization problems stated in \eqref{eqregmain2} and \eqref{eqregmain3} with efficient gradient-based optimization methods, the partial derivatives of the distance measures with respect to the transformation parameters of $T$ are required.

\subsection{Gradients}

The derivative of \eqref{eq:crisp_pts}, the crisp point-to-set distance measure $d(T(x),S)$ (in $n$-dimensional space), with respect to parameters $T_i$ of the transformation $T$ applied to a point $x \in X$, yielding $y=T(x)$,
can be written as
\begin{equation}
\frac{\partial d}{\partial T_i} = \sum_{j=1}^{n}{\frac{\partial d}{\partial y_j}\frac{\partial y_j}{\partial T_i} }.
\label{eqderiv1}
\end{equation}
The values $\frac{\partial d}{\partial y_j}$ 
are the components (partial derivatives) of the gradient $\nabla d(y, S)$ of the point-to-set distance in point $y \in Y \subset \R^n$, and are not dependent on the transformation model.

The gradient of the fuzzy point-to-set distance measure \eqref{eq:dalpha}
is given by the integral over $\alpha$-cuts, of gradients of the (crisp) point-to-set distances: 
\begin{equation}
\label{eq:dalphadiff}
\nabla d(\fuzzy{x}, \fset) = \int_0^{h(\fuzzy{x})} \nabla d(x,\alphacut{\fset}) \, \mathrm{d}\alpha\,.
\end{equation}

\subsection{Algorithms for Digital Images on Rectangular Grids}

The distances and gradients can be computed efficiently for the special case of digital images on rectangular grids.
For images quantized to $\alphalevelsconstant \in \mathcal{N}_{>0}$ non-zero discrete $\alpha$-levels the integrals in \eqref{eq:dalpha} and \eqref{eq:dalphadiff} are suitably replaced by sums. The number of quantization levels is typically taken to be a small constant; a choice of $\ell=7$ non-zero equally spaced $\alpha$-levels has previously shown to provide a good trade-off between performance, speed and noise-sensitivity \cite{lindblad2014linear}, and we keep it for all experiments.  

We need a discrete operator to approximate the gradient of $d(x,S)$ for a set $S$ defined on a rectangular grid with spacing $s \in \R_{>0}^n$. 
We propose to use the following difference operator providing a discrete approximation of  $\nabla d(x, S)$\,:
\begin{equation}
\Delta d(x) = \gamma_{x}(\delta_{1}[d](x), \ldots, \delta_{n}[d](x))\,,
\end{equation}
where 
\begin{equation}
\delta_{i}[d](x) = \textstyle\frac{1}{2 s_{i}} (d(x + s_{i}u_{i}, S) - d(x - s_{i}u_{i}, S))\,,
\end{equation}
$\gamma_{x}$ is an indicator function,
\begin{equation}
\gamma_{x} = 
\left\{\begin{array}{ll}
        1, \;\; & \text{for } \,\, d(x, S) \neq 0\\
        0, \;\; & \text{for } \,\, d(x, S) = 0\,,
        \end{array}\right.
\end{equation}
and $u_i$ is the unit-vector along the $i$-th dimension. 

The indicator function $\gamma_x$ causes the operator $\Delta[S](x)$ to be zero-valued for points included in $S$ (i.e., where the distance transform is zero-valued). This prevents discretization issues along set boundaries, where the standard central difference operator yields non-zero gradients, which can cause the measure to overshoot a potential voxel-perfect overlap.

\begin{algorithm}[H]
\caption{Distance and Gradient Maps}\label{alg:dgstack}
\begin{algorithmic}[1]
 \renewcommand{\algorithmicrequire}{\textbf{Input:}}
 \renewcommand{\algorithmicensure}{\textbf{Output:}}
 \Require Image $\fseta$. Mask $M_A$. Set of $\alpha$-levels $\set{\alpha_1,\ldots,\alpha_{\alphalevelsconstant}}$.
 \Ensure  Stack of pre-computed distance sums $D[0 \dots \alphalevelsconstant]$, and corresponding discrete gradient approximations $G[0 \dots \alphalevelsconstant]$.\smallskip
\Procedure{$\Delta\alpha\textrm{DT}$}{$\fseta, M_A, (\alpha_1, \ldots, \alpha_\alphalevelsconstant), d_\textrm{MAX}$}
 \State $\alpha_0\leftarrow 0$, $D[0] \leftarrow \mathbf{0}$, $G[0] \leftarrow (\mathbf{0}, \ldots, \mathbf{0})$
 \For{$i = 1$ to $\alphalevelsconstant$}
   \State $\textrm{DT}_i \leftarrow \textrm{DT}[\alphacutsub{\fseta}{i} \cap M_A]$\label{alg:line:distancetransform}
   \Comment{Compute DT of $\alpha$-cut.}
   \State $\textrm{DT}_i \leftarrow \min(\textrm{DT}_i, d_\textrm{MAX})$
   \ForAll{$v \in X_{\fseta}$}
     \State $D[i](v) \leftarrow (\alpha_i - \alpha_{i-1}) \textrm{DT}_i(v) + D_{i-1}(v)$
     \State $G[i](v) \leftarrow (\alpha_i - \alpha_{i-1}) \Delta \textrm{DT}_i(v) + G_{i-1}(v)$
   \EndFor
 \EndFor
 \State \textbf{return} $D, G$
 \EndProcedure
\smallskip
\Procedure{$\Delta\alpha\textrm{DT\_BD}$}{$\fseta, M_A, (\alpha_1, \ldots, \alpha_\alphalevelsconstant), d_\textrm{MAX}$}
 \State $D, G \leftarrow \Delta\alpha\textrm{DT}(\fseta, M_A, (\alpha_1,\ldots,\alpha_\alphalevelsconstant), d_\textrm{MAX})$
 \State $\setcomplement{D}, \setcomplement{G} \leftarrow \Delta\alpha\textrm{DT}(\setcomplement{\fseta}, M_A, (1-\alpha_\alphalevelsconstant,\ldots,1-\alpha_1), d_\textrm{MAX})$
 \For{$i = 0$ to $\alphalevelsconstant$}
   \ForAll{$v \in X_{\fseta}$}
     \State $D[i](v) \leftarrow D[i](v) + \setcomplement{D}[\alphalevelsconstant-i](v)$
     \State $G[i](v) \leftarrow G[i](v) + \setcomplement{G}[\alphalevelsconstant-i](v)$
   \EndFor
 \EndFor
 \State \textbf{return} $D, G$
 \EndProcedure
 \end{algorithmic}
\end{algorithm}

By creating tables for the distance and gradient sums for each image as a pre-processing step, using either of the procedures in Alg. \ref{alg:dgstack} ($\Delta\alpha\textrm{DT}$ for inwards distances and $\Delta\alpha\textrm{DT\_BD}$ for bidirectional distances), the distance and gradient may then be readily computed with Alg. \ref{alg:distance}. $\card{T}$ denotes the number of parameters of the transformation, which is $6$ for two-dimensional (2D) affine transformations, and $12$ for three-dimensional (3D) affine transformations.

\begin{algorithm}[H]
\caption{Point-to-Set Distance and its Gradient w.r.t. $T$}\label{alg:distance}
\begin{algorithmic}[1]
 \renewcommand{\algorithmicrequire}{\textbf{Input:}}
 \renewcommand{\algorithmicensure}{\textbf{Output:}}
 \Require Fuzzy point $\fseta(v)$ and associated weight $w_{A}(v)$, fuzzy set $B$ and associated (binary) mask $M_{B}$, distances $D[0 \dots \alphalevelsconstant]$ and gradients $G[0 \dots \alphalevelsconstant]$; transformation $T$.
 \Ensure  Point-to-set distance $d$, derivatives $\frac{\Delta d}{\Delta T}$, weight $w$.
\smallskip
\Procedure{D\_PTS}{$\fseta(v), \fsetb; T, w_{A}(v), M_{B}, D_{B}, G_{B}$}
     \State $\hat{v} \leftarrow T(v)$
     \If{$\hat{v} \in M_{B}$} \label{alg:line:masktest}
       \State $h \leftarrow \textrm{QUANTIZE}(\mu_\fseta(v))$
       \State $d, \mathbf{g} \leftarrow \textrm{INTERPOLATE}(D_{B}[h], G_{B}[h], \hat{v})$
       \For{$i = 1$ to $\card{T}$}
         \State $\frac{\Delta d}{\Delta T_i} \leftarrow \sum_{j=1}^{n}{\frac{\partial \hat{v}_j}{\partial T_i} \mathbf{g}_j}$           
       \EndFor
       \State \textbf{return} $w_{A}(v)d, w_{A}(v)\frac{\Delta d}{\Delta T}, w_{A}(v)$
     \Else
       \State \textbf{return} $0, \mathbf{0}, 0$
       \Comment{Zero dist., grad., and weight.}
     \EndIf
 \EndProcedure
 \end{algorithmic}
\end{algorithm}

\begin{figure}[t!]
\centering
\subfloat[Binary Image ($\alpha$-cut)\label{dttable_binimg}]{
\includegraphics[width=3.5cm,height=2.7cm]{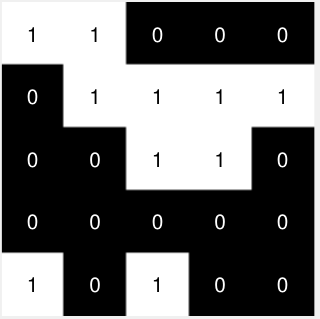}}\hspace{0.5cm}
\subfloat[Mask]{
\includegraphics[width=3.5cm,height=2.7cm]{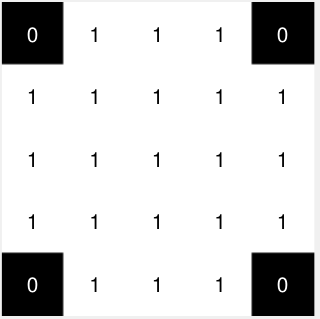}}\\
\subfloat[Masked Image ($\alpha$-cut)]{
\includegraphics[width=3.5cm,height=2.7cm]{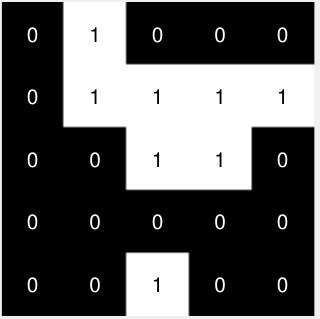}}\hspace{0.5cm}
\subfloat[Distance Transform\label{dttable_dtimg}]{
\includegraphics[width=3.5cm,height=2.7cm]{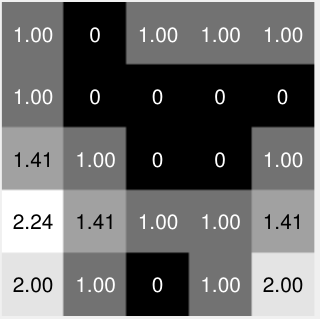}}\\
\subfloat[Horizontal Gradient\label{dttable_gradximg}]{
\includegraphics[width=3.5cm,height=2.7cm]{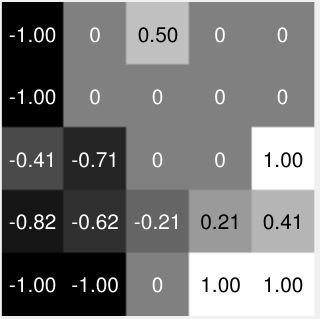}}\hspace{0.5cm}
\subfloat[Vertical Gradient\label{dttable_gradyimg}]{
\includegraphics[width=3.5cm,height=2.7cm]{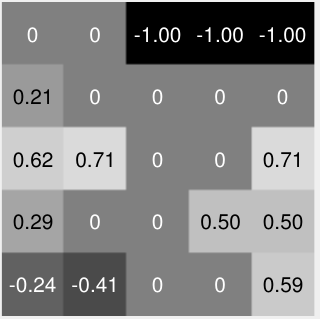}}\\
\caption{(a) Example $\alpha$-cut in a small 2D image. (b) Binary mask. (c) $\alpha$-cut after masking. (d) (Euclidean) Distance transform (for (c)). (e-f) Gradient approximation of the distance transform $\Delta \textrm{DT}$.\label{dttable}}
\end{figure}

The procedures in Alg. \ref{alg:dgstack} have linear run-time complexity $\mathcal{O}((\alphalevelsconstant+1)\card{X_{\fseta}})$, achieved by using a linear-time algorithm for computing the distance transform (e.g. \cite{maurer2003linear}) in line \ref{alg:line:distancetransform} of Alg. \ref{alg:dgstack}. The space complexity of the algorithm is $\mathcal{O}((\alphalevelsconstant+1)\card{X_{\fseta}})$ and the $D,G$ structures must remain in memory to enable fast lookup in Alg. \ref{alg:distance}. Figure~\ref{dttable} shows an example of the distance and gradient of a sample $\alpha$-level. Alg. \ref{alg:distance} computes the point-to-set distance and gradient w.r.t. the transformation using the pre-computed tables and has run-time complexity $\mathcal{O}(\card{T}n)$ thus being independent of $\alphalevelsconstant$ and the size of $\fseta$ and $\fsetb$.

\newlength\myindent
\setlength\myindent{2em}
\newcommand\bindent{%
  \begingroup
  \setlength{\itemindent}{\myindent}
  \addtolength{\algorithmicindent}{\myindent}
}
\newcommand\eindent{\endgroup}

\begin{algorithm}[H]
\caption{Symmetric Registration}\label{alg:reg}
\begin{algorithmic}[1]
 \renewcommand{\algorithmicrequire}{\textbf{Input:}}
 \renewcommand{\algorithmicensure}{\textbf{Output:}}
 \Require Fuzzy sets $\fseta, \fsetb$, with binary masks $M_{A}, M_{B}$, and weight functions $w_{A}, w_{B}$, initial transformation guess $T$. No. of $\alpha$-levels $\ell$, distance saturation $d_{\textrm{MAX}}$, step-lengths $\lambda[{1 \ldots N}]$, and iteration count $N$ are hyper-parameters.
 \Ensure Symmetric set distance $d$ ($\dalphabidirsamdr$), estimated transformation $\hat{T}$.
  \smallskip
   \State $D_{A}, G_{A} \leftarrow \Delta\alpha\textrm{DT\_BD}(\fseta, M_A, (\alpha_1, \ldots, \alpha_\alphalevelsconstant), d_{\textrm{MAX}})$
   \State $D_{B}, G_{B} \leftarrow \Delta\alpha\textrm{DT\_BD}(\fsetb, M_B, (\alpha_1, \ldots, \alpha_\alphalevelsconstant), d_{\textrm{MAX}})$
   \For{$i=1$ to $N$}
     \State $d_{1}, \frac{\Delta d_{1}}{\Delta T}, w_{1} \leftarrow \ldots$
     \Statex \hspace{1cm} $\sum\limits_{v \in X_{\fseta}}^{}\!{  \textrm{D\_PTS}(\fseta(v), \fsetb; T, w_{A}(v), M_{B}, D_{B}, G_{B})  }$
     \State $d_{2}, \frac{\Delta d_{2}}{\Delta T^{-1}}, w_{2} \leftarrow \ldots$
     \Statex \hspace{1cm} $\sum\limits_{v \in X_{\fsetb}}^{}\!{  \textrm{D\_PTS}(\fsetb(v), \fseta; T^{-1}, w_{B}(v), M_{A}, D_{A}, G_{A})}$
     \State $T \leftarrow T - \frac{\lambda[i]}{2}\big(\frac{1}{w_1}\frac{\Delta d_{1}}{\Delta T} + \frac{1}{w_2}\frac{\Delta T^{-1}}{\Delta T}\frac{\Delta d_{2}}{\Delta T^{-1}}\big)$ \label{alg:reg:lineupd}
   \EndFor
   \State $d \leftarrow \frac{1}{2}(\frac{1}{w_1}d_{1} + \frac{1}{w_2}d_{2})$
   \Comment{Output final distance value.}
   \State $\hat{T} \leftarrow T$
   \Comment{Output final transformation estimate.}
 \end{algorithmic}
\end{algorithm}

Algorithm \ref{alg:reg} performs a complete registration given two images, their binary masks, weight functions, and an initial transformation. Algorithm \ref{alg:reg} completes $N$ full iterations, however other termination criteria may be beneficial (see Sec. \ref{sec:Impl}). The registration consists of pre-processing, followed by a loop where the symmetric set distance and derivatives are computed and $T$ is updated.
$\frac{\Delta T^{-1}}{\Delta T}$, in line \ref{alg:reg:lineupd} of Alg. \ref{alg:reg} denotes a matrix $\big[\frac{ \partial T^{-1}_j } { \partial T_i }\big]$ of partial derivatives of the parameters of the inverse transform $T^{-1}$ w.r.t. the parameters of the forward transform $T$. This matrix relates the computed partial derivatives $\frac{\Delta d_{2}}{\Delta T^{-1}}$ with the parametrization of the forward transform.

The overall run-time complexity is $\mathcal{O}(N\card{T}n(\card{X_{\fseta}} + \card{X_{\fsetb}}) + (\alphalevelsconstant+1)(\card{X_{\fseta}} + \card{X_{\fsetb}}))$. Practical choices for $N$ tend to be in the range $[1000,3000]$, depending on hyper-parameters (e.g. $\lambda$), and distance in parameter-space between starting position and the global optimum. The evaluation in Sec. \ref{sec:evaluation} confirms empirically that convergence, according to \eqref{eq:criterium:gmt} or \eqref{eq:criterium:msl}, tends to be reached after $1000$ to $3000$ iterations, using an optimizer with a decaying $\lambda$.

The $\textrm{QUANTIZE}$ procedure in Alg. \ref{alg:distance} takes the membership of point $v$, $\mu_\fseta(v)$, and gives the index $i$ of the minimal $\alpha$-level ($\alpha_1$, \dots, $\alpha_\alphalevelsconstant$) for which $\mu_\fseta(v) \geq \alpha_i$. If the membership is below all $\alpha$-levels, the index is $0$. For $\alphalevelsconstant$ equally spaced $\alpha$-levels, the quantization can be expressed as
\begin{equation}
\textrm{QUANTIZE}(\mu_\fseta(v)) = \lfloor \alphalevelsconstant \mu_\fseta(v) + 0.5 \rfloor.
\end{equation}

The $\textrm{INTERPOLATE}$ procedure in Alg. \ref{alg:distance} computes the value of the discrete functions $D$ and $G$ in point $\hat{v}$ which may not be on the grid due to application of $T$. There are many interpolation schemes proposed in the literature, but we suggest that either nearest neighbor (for maximal speed) or linear interpolation (for higher accuracy) are used here, since the distance and gradient fields are smooth.
By linearity of integration and summation, nearest neighbor and linear interpolation may be performed on the pre-processed $D$ and $G$ and yield the same result as if each level was interpolated before integration, allowing efficient computation. The (discretized) measure does not require intensity interpolation; the interpolation operates on distances and gradients only.

\section{Implementation}\label{sec:Impl}

We implemented the proposed distance measure and registration method in the open-source Insight Segmentation and Registration Toolkit (ITK)~\cite{avants2014insight}. We chose this particular software framework because it
\begin{itemize}
\item enables the use of an existing optimization framework,
\item allows for a fair comparison against well written, tested, and widely used implementations of reference similarity measures, with support for resolution-pyramids,
\item supports anisotropic/scaled voxels in relevant algorithms,
\item facilitates reproducible evaluation,
\item makes the proposed measure easily accessible for others.
\end{itemize}
The built-in ITK optimizer we have used for the registration tools and all the evaluation is {\emph{RegularStepGradientDescentOptimizerv4}}. This is an optimizer based on gradient descent, with an initial step-length $\lambda$, and a relaxation factor which reduces the used step-length gradually as the direction of the gradient changes, in order to enable convergence with high accuracy. In addition to a maximum number of iterations $N$, two termination criteria are used: (\romlcase{1}) a gradient magnitude threshold (GMT),
\begin{equation}
\label{eq:criterium:gmt}
    \sqrt{\textstyle\frac{\partial d}{\partial T_1}^2 + \ldots + \frac{\partial d}{\partial T_{\card{T}\!}}^2} < \text{GMT},
 \end{equation}
and (\romlcase{2}) a minimum step-length (MSL),
\begin{equation}
\label{eq:criterium:msl}
    \lambda r < \text{MSL},
\end{equation}
where $r$ is the current relaxation coefficient. We use default values of $0.0001$ for both of these criteria. A relaxation factor of $0.99$ is used for all experiments, since it performed well in preliminary tests; in this study we are willing to trade some (potential) additional iterations for better robustness. To maximize utilization of the limited number of $\alpha$-levels, images are normalized before registration to make sure that they are within the valid $\left[0, 1\right]$ interval. We use the following robust normalization technique: 
Let $P_{\rho}(X)$ denote the $\rho$-th percentile of image $X$ with respect to image intensities, 
\begin{equation}
    \textrm{NORM}_\rho(X) = \max\Big[0, \min\big[1, \frac{(X - P_{\rho}(X))}{P_{1-\rho}(X)- P_{\rho}(X)}\big]\Big]\,.
\end{equation}

\section{Performance Analysis}
\label{sec:evaluation}

We evaluate performance of the proposed method, both for 2D and 3D images, in two different scenarios; (\romlcase{1}) we perform a statistical study on synthetically generated images, where we seek to recover known transformations and measure the registration error by comparing the ground truth locations of known landmarks with the corresponding registered ones; (\romlcase{2}) we apply the proposed framework to real image analysis tasks: landmark-based evaluation of registration of TEM images in 2D, and atlas-based segmentation evaluation of 3D MR images of brain.

To compare the proposed measure and registration method against the most commonly used alternative methods and similarity measures, we select the widely used ITK implementations of optimization framework and similarity measures (SSD, PCC and MI) as the baseline of intensity-based registration accuracy. Note that the PCC measure is denoted Normalized Cross Correlation (NCC) in the ITK framework.

All experiments are performed on a workstation with a 6-core Intel i7-6800K, 3.4GHz processor with 48GB of RAM and 15MB cache. The operating system used is Ubuntu 16.04 LTS. The compiler used to build the framework is g++ version 5.4.0 (20160609). Version 4.9 of the ITK-framework is used for testing and evaluation.

\subsection{Datasets}

One biomedical 2D dataset and one medical 3D dataset are used for the evaluation.

\subsubsection{TEM Images of Cilia (2D)}

The dataset of 20 images of cilia \cite{suveer2017enhancement} 
is acquired with the MiniTEM\footnote{MiniTEM imaging system is developed by Vironova AB.} imaging system.  Each image is isotropic of size $129 \times 129$ pixels, with a pixel-size of a few \textit{nm}. An example is shown in Fig.~\ref{fig:featurefail}. A particular challenge is the near-rotational symmetry of the object: 9 pairs of rings are located around a central pair of rings, which gives 9 plausible solutions for a registration problem. The alignment of the central pair can be taken into special consideration to reduce the number of solutions to two. The dataset comes with a set of 20 landmarks per image, indicating the position of each of the relevant structures to be detected and analysed --  $20$ rings ($2$ in the center and $18$ in a circle around the center). The landmarks are produced by a domain expert and are only used for evaluation of the registrations.

\subsubsection{LPBA40 (3D)}

LPBA40 \cite{shattuck2008construction} is a publicly available dataset of 40 3D images of brains of a diverse set of healthy individuals, acquired with MRI. The images are anisotropic, of size $256 \times 124 \times 256$\,voxels with voxel-size $0.86 \times 1.5 \times 0.86\,\textit{mm}^3$. The dataset comes with segmentations of the brains into 56 distinct regions marked by a medical expert, which are used in this study as ground-truth for evaluation. LPBA40 includes two atlases: 
first 20 out of 40 MR images of brain in the dataset are used to generate one brain atlas by
\emph{Symmetric Groupwise Normalization} (SyGN) \cite{avants2010optimal}; another atlas is created analogously, from the last $20$ brains in the dataset. The atlases contain both a synthesized MR image and the fused label category in all the voxels, as well as a whole brain mask which may be used for brain extraction.

\subsection{Evaluation Criteria}

We evaluate accuracy and robustness of the registration methods in presence of noise,  their robustness w.r.t. change of roles of reference and floating image (symmetry), and their speed. We quantify the performance of the observed frameworks in terms of the following quality measures:  

\subsubsection{Average Error measure (AE)}
The registration result is quantified as the mean Euclidean distance between the sets of corresponding image corner landmarks $L_{\text{R}}$ and $T(L_{\text{F}})$ in the reference image space, after transformation of the floating image corner landmarks $L_{\text{F}}$, where $\card{L_{\text{R}}}$ is the number of landmarks (4 in 2D; 8 in 3D). The quality measure is defined as
\vspace*{-3mm}\begin{equation}
\avgerr{(T; L_{\text{R}}, L_{\text{F}})} = \frac{1}{|L_{\text{R}}|}\sum_{i = 1}^{|L_{\text{R}}|}{\deuclidean(L_{\text{R}}(i), T(L_{\text{F}}(i)))}.
\end{equation}

A slight variation of this measure, the Average Minimal Error (AME), is used in the real task of cilia registration: 
\begin{equation}
\avgminerr{(T; L_{\text{R}}, L_{\text{F}})} = \frac{1}{|L_{\text{R}}|}
\sum_{i = 1}^{|L_{\text{R}}|}{ \min\limits_{x \in L_{\text{F}}} {\deuclidean(L_R(i), T(x))} }\,.
\end{equation}
For the central pair, the error is simply $\avgminerr_{\textrm{CP}} = \avgminerr$, whereas for the 
outer rings we utilize the knowledge that an odd (even) landmark should be matched with an odd (even) landmark of the other image. 
The error function for the outer rings, \cite{suveer2017enhancement}, is therefore defined as:
\begin{equation}
\label{cilia_ring_matching}
\begin{split}
\avgminerr_{\textrm{Outer}}{(T; L^\text{Odd}_{\text{R}}, L^\text{Odd}_{\text{F}}, L^\text{Even}_{\text{R}}, L^\text{Even}_{\text{F}})} = \\ {\textstyle\frac{1}{2}}(\avgminerr(T; L^\text{Odd}_{\text{R}}, L^\text{Odd}_{\text{F}})\!+\!\avgminerr(T; L^\text{Even}_{\text{R}}, L^\text{Even}_{\text{F}}))\,.
\end{split}
\end{equation}

\subsubsection{Success Rate (SR)}
A registration is considered successful if its AE is below one voxel(pixel).
Success rate (SR) 
 at a given AE value corresponds to the ratio of successful registrations (w.r.t. the set of performed ones).  

\subsubsection{Symmetric Success Rate (SymSR)}
is defined as the ratio of performed registrations which are successful (i.e., $AE \leq 1$) in both directions, i.e., when the roles of reference and floating image are exchanged. 

\subsubsection{Inverse Consistency Error (ICE),\cite{christensen2001consistent}}

Given a set
of interest $X_A \subseteq A$, the transformations  $T_{\textrm{AB}} \colon A \to B$, and $T_{\textrm{BA}} \colon B \to A$, the ICE of this pair of transformations is
\begin{equation}
\textrm{ICE}(T_{\textrm{AB}}, T_{\textrm{BA}}; X_A) = \frac{1}{\card{X_A}} \sum\limits_{x \in X_A}^{}{\deuclidean(T_{\textrm{BA}}(T_{\textrm{AB}}(x)), x)}.
\end{equation}
We compute ICE considering all the points of the reference image for each of the cases where Symmetric Success is observed ($AE \leq 1$ in both directions).

\subsubsection{Jaccard Index for segmentation evaluation} 

For two binary sets, $R_1$ and $R_2$, the Jaccard Index
is defined as
\begin{equation}
J(R_1, R_2) = \frac{\setcardinality{(R_1 \cap R_2)}}{\setcardinality{(R_1 \cup R_2)}}.
\end{equation}

\subsubsection{Execution Time}

 We evaluate (\romlcase{1}) the execution times required for one iteration in the registration procedure, i.e.,   times needed to  compute the distance (similarity) measure and its derivatives, with full sampling, and in full image resolution, 
between two distinct images from the same set,
as well as (\romlcase{2}) the execution time for complete registrations.

\subsection{Parameter Tuning}

The distance measure and optimization method have a number of parameters which must be properly chosen. Synthetic tests indicated that the following values lead to good optimization performance: three pyramid levels with downsampling factors $(4, 2, 1)$  and Gaussian smoothing $\sigma = \left(5.0, 3.0, 0.0\right)$,  max $3000$ iterations per level and an initial step-length $\lambda=0.5$. 
The number of $\alpha$-levels used is $\alphalevelsconstant = 7$, which has shown to provide a reasonable trade-off between computational costs, sensitivity to significant variations in intensity and robustness to noise \cite{lindblad2014linear}. The optimal value for $\alphalevelsconstant$ is application-dependent; in essentially all observed cases, $\alphalevelsconstant > 1$ (non-crisp) outperforms a crisp (binarized) representation.
Normalization percentile is normally $5\%$. This same parameter setting, if not stated differently, is used in all the tests, on both synthetic and real data.

\subsection{Synthetic Tests}

A synthetic evaluation framework is used to 
evaluate the performance of the proposed method, and to compare it with standard tools based on SSD, PCC, and MI, in a controlled environment. 
 For this evaluation, we construct sets of transformed versions of a reference image and add (a new instance of) Gaussian noise to each generated image. The transformations are selected at random from a multivariate uniform distribution of rotations measured in degrees (1 angle for 2D images and 3 Euler angles for 3D images) and translations measured in fractions of the original image size.

\begin{figure*}[t]
\centering
\subfloat[Reference image.\label{ciliarefimage}]{\hspace{1cm}\includegraphics[width=2.1cm]{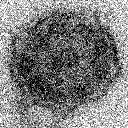}\hspace{1cm}}
\subfloat[Reference mask.\label{ciliarefmask}]{\hspace{1cm}\includegraphics[width=2.1cm]{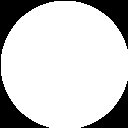}\hspace{1cm}}
\subfloat[Floating image.\label{ciliafloatimage}]{\hspace{1cm}\includegraphics[width=2.1cm]{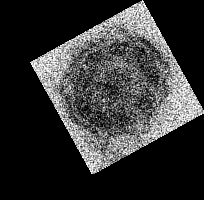}\hspace{1cm}}
\subfloat[Floating mask.\label{ciliafloatmask}]{\hspace{1cm}\includegraphics[width=2.1cm]{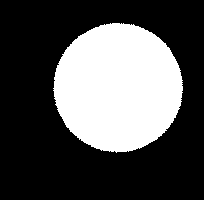}\hspace{1cm}}

\subfloat[Small transformations]{\includegraphics[width=6.0cm]{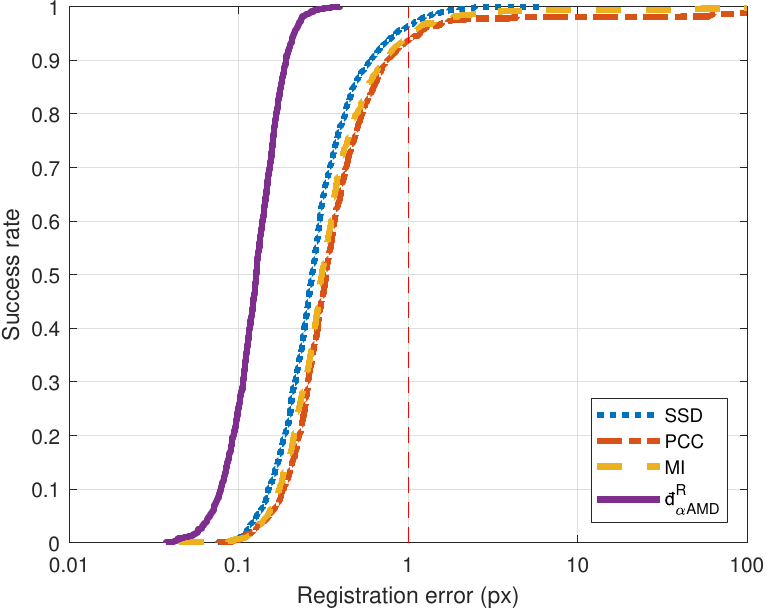}} 
\subfloat[Medium transformations]{\includegraphics[width=6.0cm]{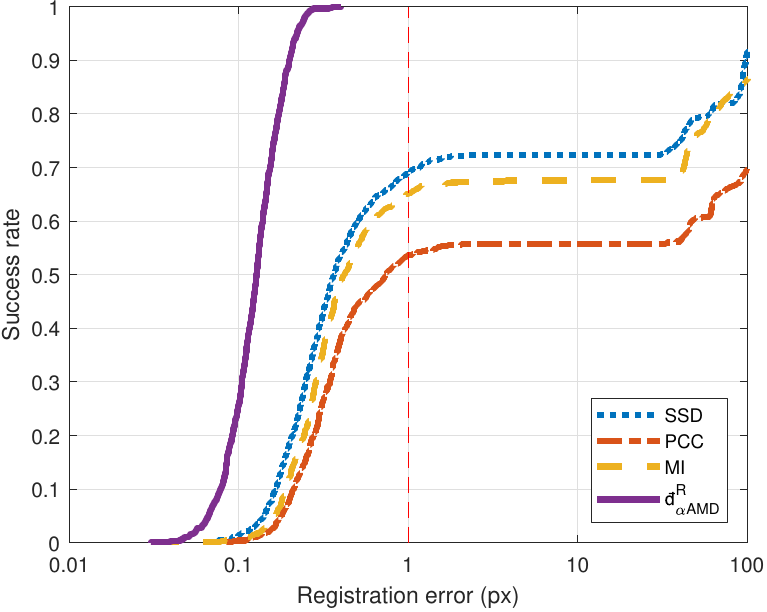}}
\subfloat[Large transformations]{\includegraphics[width=6.0cm]{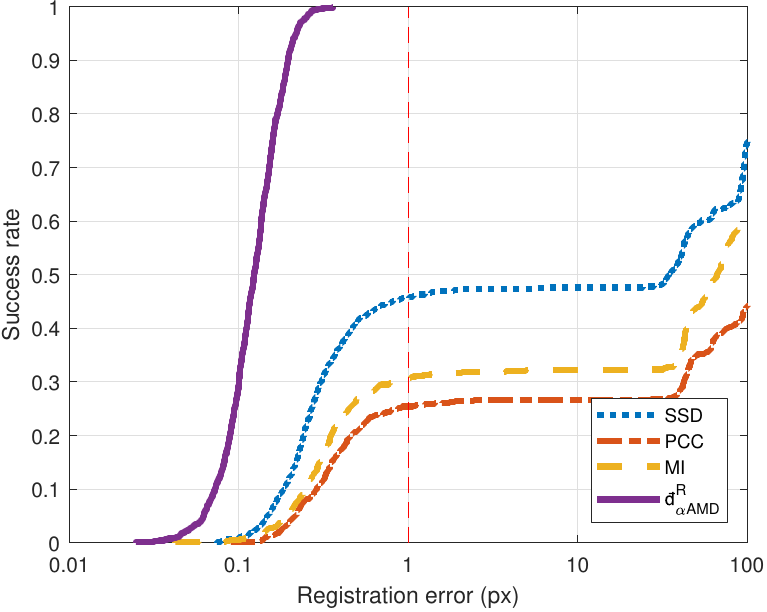}} \\
\caption{Registration error for 2D TEM images of cilia with Gaussian noise of $\sigma=0.1$ added, for three observed transformation classes. (a-d) Examples of reference-floating image pair with corresponding masks. (e-g) Cumulative histograms of the fraction of registrations with registration error AE below a given value (left and up is better). The red vertical line shows the chosen threshold for success, $AE\leq 1$.}
\label{fig:ciliasynthresult}
\end{figure*}

\subsubsection{2D TEM Images of Cilia}

Three sets of transformed images are 
built based on image Nr. 1 in the observed dataset, by applying on it the following three groups of transformations:  {\it Small},  containing compositions of translations of up to $10\%$ of image size (in any direction) and rotations by up to $10\degree$; {\it Medium}, containing compositions of translations and rotations such that at least one of the parameters exceeds the range of Small, and falls within $10-20\%$ of image size of translation (in at least one direction), or $10-20\degree$ of rotation; and {\it Large}, containing compositions of translations and rotations such that at least one of the parameters exceeds the range of Medium, and falls within $20-30\%$ of image size of translation (in at least one direction), or $20-30\degree$ of rotation. The transformed images are also corrupted by additive Gaussian noise, from $\mathcal{N}(0, 0.1^2)$ ($\sigma\!=\!0.1$, corresponding to a PSNR$\,\approx\,$20\,dB). 
Each group of transformations is applied 1000 times, and the resulting images are registered to image Nr. 1, each time corrupted by a new instance of Gaussian noise.   

To evaluate symmetry, we performed $1000$ registrations of images transformed by randomly selected translations of up to $30\%$ of image size, and rotations by up to $30\degree$, and corrupted by additive noise from $\mathcal{N}(0, 0.1^2)$. Each of the registrations were performed twice, with exchanged roles of reference image and floating image. 

Intensity-based registration with gradient-descent optimization can be computationally demanding, requiring the distance function and its derivative for each iteration of the optimization procedure. The time to compute the distance and derivatives is directly proportional to the number of sampled points. We, therefore, evaluate influence of the sampling fraction on registration success, observing registrations after {\it Small} transformations and added noise (with $\sigma=0.1$), over a range of sampling fractions.  For each evaluated sampling fraction, $1000$ registrations are performed and SR and AE are computed for successful registrations ($AE \leq 1$). No resolution pyramids are used for these tests.

\subsubsection{3D MR Images of Brain}
Three sets of transformed images are 
built based on image Nr. 1 in the observed dataset, by applying to it the following three groups of transformations: 
 {\it Small},  containing compositions of translations of up to $10\%$ of image size (in any direction) and rotations by up to $10\degree$ (around each of the rotation axes); {\it Medium}, containing compositions of translations and rotations such that at least one of the parameters exceeds the range of Small, and falls within $10-15\%$ of image size of translation (in at least one direction), or $10-15\degree$ of rotation (around at least one rotation axes); and {\it Large}, containing compositions of translations and rotations such that at least one of the parameters exceeds the range of Medium, and falls within $15-20\%$ of image size of translation (in at least one direction), or $15-20\degree$ of rotation (around at least one rotation axes).
 The transformed images are also corrupted by additive Gaussian noise, from $\mathcal{N}(0, 0.1^2)$. 
Each group of transformations is applied 200 times, and the resulting images are registered to image Nr. 1, each time corrupted by a new instance of Gaussian noise. 

\subsection{Results of Synthetic Tests} 

\subsubsection{2D TEM Images of Cilia}
Figure~\ref{fig:ciliasynthresult} shows the distributions of registration errors (AE), for the three transformation classes. Superiority of the proposed measure, and the corresponding registration framework, is particularly clear for Medium and Large transformations; it reaches a $100\%$ success rate, with subpixel accuracy, whereas the competitors not only exhibit considerably lower accuracy, but also much lower success rate, i.e., they completely fail in a large number of cases.

\begin{table}[tb]
    \caption{Registration of synthetic 2D images of cilia. The tables show success rate (SR), average error (AE) of successful registrations, symmetric success rate (SymSR), average inverse consistency error (ICE) and average runtime for the registration with complete sampling (a) and with random subsampling (b). Successful registrations ($AE \leq 1$) of transformations up to (and including) Large, are considered. }
    \label{tab:ciliaalltable}
   \subfloat[Full sampling]{
    \begin{tabular}{c||c|c|c|c|c}
         Measure & SR & AE & SymSR & ICE & Time (s) \\ \hline
         SSD & 0.536 & 0.3086 & 0.313 & 0.2424 & 17.3 \\
         PCC & 0.363 & 0.3413 & 0.249 & 0.4227 & 20.8 \\
         MI & 0.440 & 0.3495 & 0.251 & 0.4518 & 18.3\\
         $\dalphabidirsamdr$ & $\mathbf{1.000}$ & $\mathbf{0.1295}$ & $\mathbf{1.000}$ & $\mathbf{0.0023}$ & $\mathbf{3.49}$ \\
    \end{tabular}} \\
    \subfloat[0.1 sampling fraction]{
    \begin{tabular}{c||c|c|c|c|c}
         Measure & SR & AE & SymSR & ICE & Time (s) \\ \hline
         SSD & 0.367 & 0.6270 & 0.186 & 0.5260 & 1.761 \\
         PCC & 0.299 & 0.6364 & 0.152 & 0.5676 & 2.171 \\
         MI & 0.283 & 0.6219 & 0.068 & 0.5974 & 2.083 \\
         $\dalphabidirssamdr$ & $\mathbf{1.000}$ & $\mathbf{0.1311}$ & $\mathbf{1.000}$ & $\mathbf{0.0193}$ & $\mathbf{0.834}$ \\
    \end{tabular}} 
\end{table}

\begin{figure}[b]
\centering
\includegraphics[width=\columnwidth]{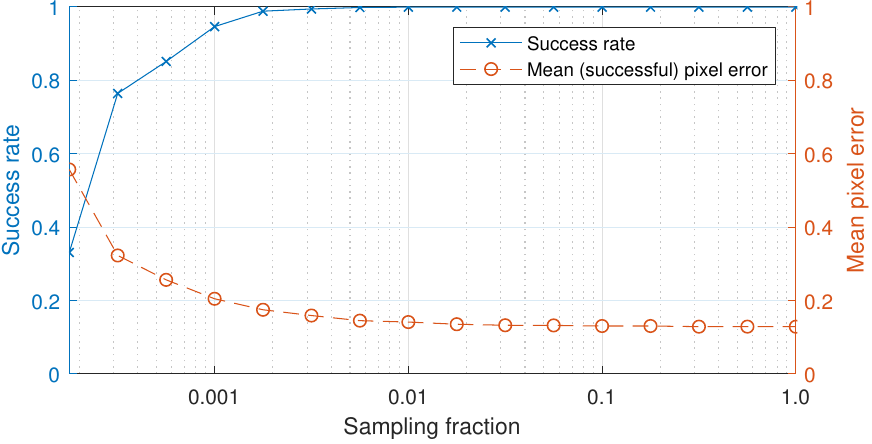}
\caption{(Left/Blue) SR for registrations of cilia images, and (Right/Red) AE of the successful registrations, as functions of sampling fraction for the proposed method. Both measures improve (almost) monotonically with sampling fraction and flatten out after approximately $0.01$.}
\label{fig:samplingfraction}
\end{figure}

\begin{figure*}[t]
\centering
\subfloat[Reference (XY)\label{refbrainxyimage}]{\hspace{0.4cm}\includegraphics[width=2.2cm, height=2.2cm]{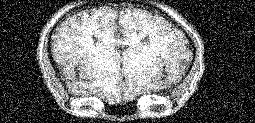}\hspace{0.4cm}}
\subfloat[Reference (XZ)\label{refbrainxzimage}]{\hspace{0.4cm}\includegraphics[width=2.2cm, height=2.2cm]{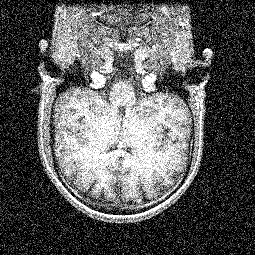}\hspace{0.4cm}}
\subfloat[Reference (YZ)\label{refbrainyzimage}]{\hspace{0.4cm}\includegraphics[width=2.2cm, height=2.2cm]{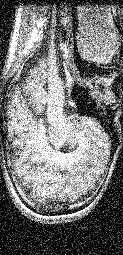}\hspace{0.4cm}}
\subfloat[Floating (XY)\label{floatxyimage}]{\hspace{0.4cm}\includegraphics[width=2.2cm, height=2.2cm]{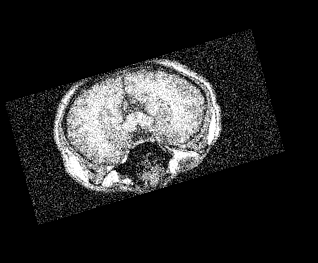}\hspace{0.4cm}}
\subfloat[Floating (XZ)\label{floatxzimage}]{\hspace{0.4cm}\includegraphics[width=2.2cm, height=2.2cm]{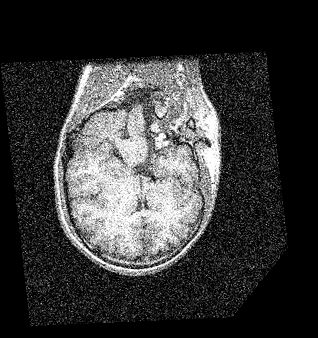}\hspace{0.4cm}}
\subfloat[Floating (YZ)\label{floatyzimage}]{\hspace{0.4cm}\includegraphics[width=2.2cm, height=2.2cm]{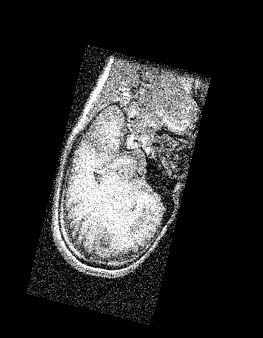}\hspace{0.4cm}}

\subfloat[Small transformations]{\includegraphics[width=6.0cm]{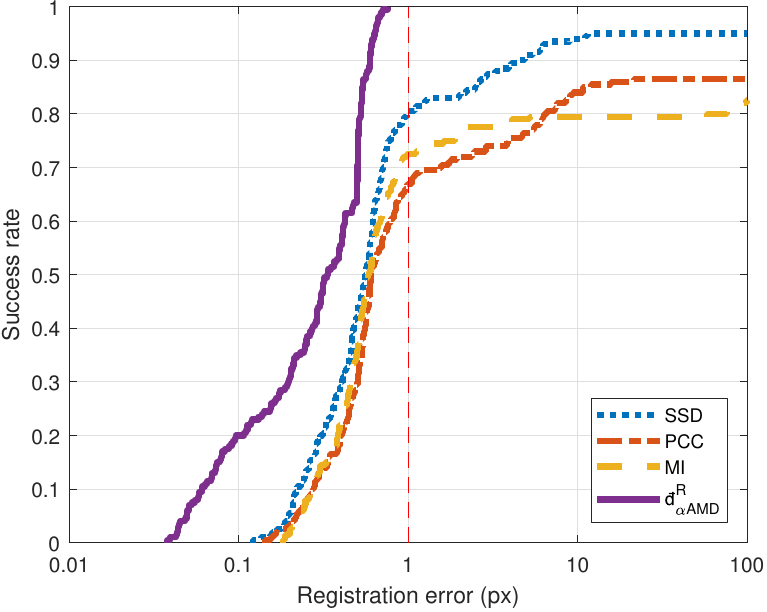}}
\subfloat[Medium transformations]{\includegraphics[width=6.0cm]{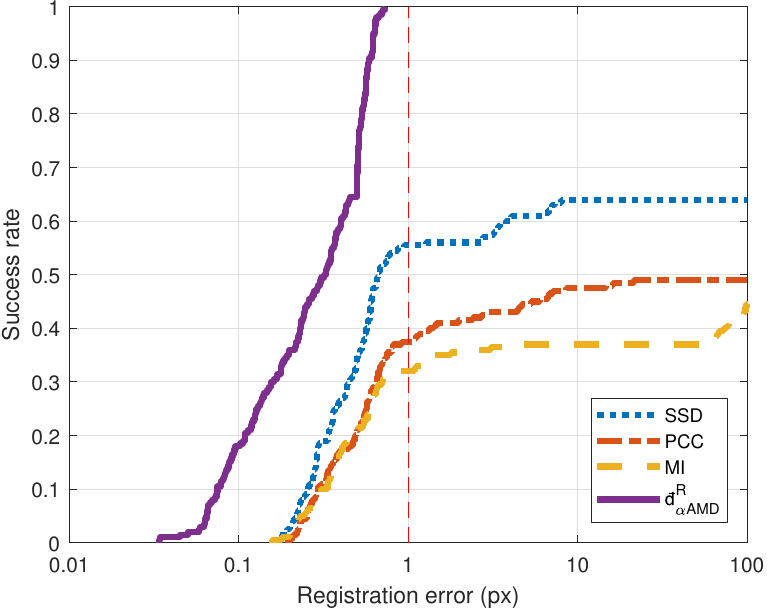}} 
\subfloat[Large transformations]{\includegraphics[width=6.0cm]{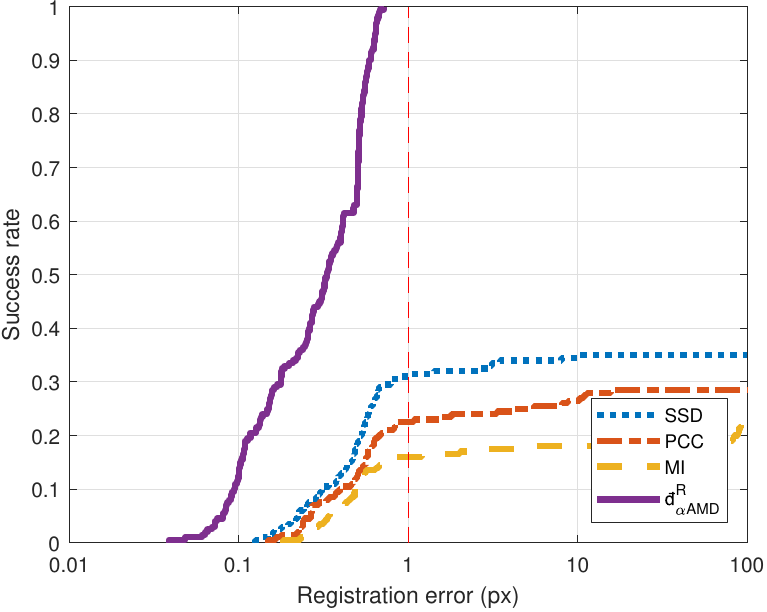}}\\

\caption{Registration error for 3D MR images of brain with Gaussian noise, $\sigma=0.1$, added, for three observed transformation magnitudes classes. (a-f) Example of reference-floating image pair in slices along each major axis. (g-i) Cumulative histograms of the fraction of registrations with registration error AE below a given value (left and up is better). The red vertical line shows the chosen threshold of success, $AE\leq 1$.}
\label{fig:lpba40synthresult}
\end{figure*}

\begin{figure*}[t]
\centering
\subfloat[Success-rate (\%)]{\includegraphics[width=5.7cm]{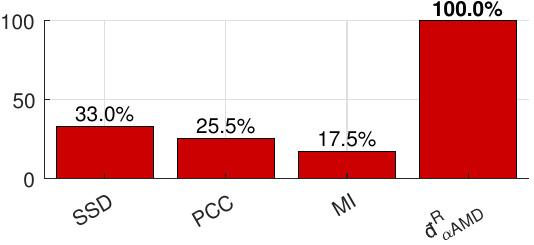}}\hfill
\subfloat[Mean error (px)]{\includegraphics[width=5.7cm]{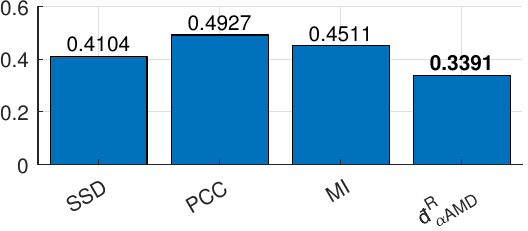}}\hfill
\subfloat[Execution time (s)]{\includegraphics[width=5.7cm]{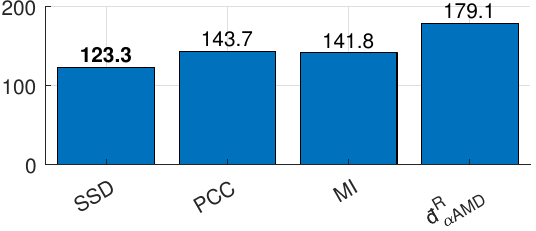}}
\caption{Results of synthetic registration of 3D brain images from the LPBA40 dataset. The plots show the (a) success-rate (SR), (b) mean error (ME) for successful registrations and (c) the average runtime in seconds for the registration with random subsampling with 0.01 sampling fraction. (a) Higher is better. (b-c) Lower is better. Bold marks the best result w.r.t. each statistic.}
\label{fig:lpba40barplots}
\end{figure*}

Overall registration performance is summarized in Table~\ref{tab:ciliaalltable}, for complete sampling (a), and for random sampling of 10\% of the points (b).
The proposed method has $100\%$ success rate and also $100\%$ symmetric success rate. The other observed measures exhibit much lower success rate and poor symmetry scores; the second best, SSD, succeeds in $54\%$ of the cases, and succeeds symmetrically in only $31\%$ of the cases. The registration error for successful registrations is considerably smaller for the proposed method, while the execution time is considerably lower. The reduced sampling fraction in (b) has a small impact on the proposed method while substantially degrading the performance of the other measures.

Figure~\ref{fig:samplingfraction} shows registration performance for varying sampling fractions;  Small transformations, in presence of noise ($\sigma=0.1$) are considered. We observe that the registration performance flattens and stabilizes at approximately $0.01$ sampling fraction ($1\%$ of the points).
We conclude that previous findings of \cite{viola1997alignment,klein2007evaluation}, suggesting that random subsampling provides good performance even with very small sampling fractions, apply well for the proposed measure.

\subsubsection{3D MR Images of Brain}
Figure~\ref{fig:lpba40synthresult} shows the observed distributions of registration errors (AE) for the three transformation classes, and clearly confirms that the proposed method is robust and with high performance, even for larger transformations, while the magnitude of the transformation has a substantial negative effect on the performance of the other observed measures.

Figure~\ref{fig:lpba40barplots} presents bar plots corresponding to the performed synthetic tests on the LPBA40-dataset, consisting of $200$ registrations of images after up to (and including) Large transformations (with additive Gaussian noise, $\mathcal{N}(0,0.1^2)$). Successful registrations ($AE \leq 1$) are observed. Here as well, the proposed method delivers $100\%$ success rate, whereas the second best, SSD, succeeds in only $33\%$ of the cases. The registration error for successful registrations is the smallest for the proposed method. We observe a relative increase in execution time of the proposed registration framework in 3D case, where it is slightly slower than the other measures. 

\subsubsection{Execution Time Analysis}

The number of iterations required for convergence of the optimization (registration) typically range from 1000 to 3000. 
Measures SSD, PCC and MI use cubic spline interpolation. Lookups from the distance maps for $\dalphabidirsamdr$ are done using linear interpolation. Table \ref{timeanalysistable} shows the mean (and standard deviation) execution time of one iteration, which includes computation of the measures and their derivatives, repeated $1000$ times for 2D, and $50$ times for 3D affine image registrations. We observe that the proposed measure is the fastest per iteration both in 2D and 3D. Note that these execution time measurements exclude pre-processing.

\begin{table}[t]
\caption{Time analysis of distance (similarity) value and derivative computations for a full resolution image, repeated to generate statistics. Bold marks the fastest measure in each category (2D and 3D). The 2D images are of size $1600 \times 1278$, and the 3D images are of size $256 \times 124 \times 256$.}
\label{timeanalysistable}
\centering
\begin{tabular}{l|cc|cc}
Measure & \multicolumn{2}{c|}{Cilia (2D) [s]} & \multicolumn{2}{c}{Brain (3D) [s]} \\
 &  Mean & Std.dev. & Mean & Std.dev. \\ \hline
 $\dalphabidirsamdr$ & \textbf{0.270} & 0.010 & \textbf{3.116} & 0.098\\
 SSD & 0.718 & 0.036 & 4.782 & 0.066 \\
 PCC & 1.191 & 0.026 & 8.562 & 0.002 \\
 MI & 0.890 & 0.025 & 5.699 & 0.002 \\
\end{tabular}
\end{table}

\subsection{Evaluation on Real Applications}

\subsubsection{Registration of Cilia}
\label{sect:ciliaregistration}

Registration of multiple cilia instances detected in a single TEM sample, for enhancement of diagnostically relevant sub-structures, requires a pixel-accurate and robust method which is able to overcome the challenges posed by the near-rotational symmetry of a cilium. 
At most two of the possible solutions properly align the central pair, which is vital for a successful reconstruction.

We compare the performance of the proposed method with reported results of a previous study \cite{suveer2017enhancement} which uses intensity-based registration with PCC as similarity measure. We follow the general protocol described in \cite{suveer2017enhancement} and perform, as a first step, a multi-start rigid registration (parameterized by angle $\theta$ in radians, and translation $\mathbf{t} = (t_x, t_y)$), followed, in a second step, by affine registration initiated by the best (lowest final distance) registration of the 9 rigid ones.  

No resolution pyramids are used since they were observed to interfere with the multi-start approach (by facilitating large movements). The registrations are performed in full resolution, without stochastic subsampling. For the rigid registration we use a small circular binary mask with radius of  $24$ pixels, positioned in the center, combined with a squared circular Hann window function. The affine registration is performed using a  circular binary mask with radius of $52$ pixels; the mask removes the outside background and the outer plasma which is not helpful in guiding the registration. No additional weight-mask is used for the affine registration. Step length $0.1$ was used for the rigid and $0.5$ for the affine registration. We use $\alphalevelsconstant = 7$. Normalization percentile is set to $0\%$ for the rigid stage and $1\%$ for the affine stage.

A feature-based approach is also included in this performance evaluation. The SIFT feature-detector \cite{lowe1999object}, with RANSAC \cite{fischler1981random} as model fitting and correspondence point filtering method, as implemented in FIJI, is evaluated with both rigid and affine transformation models. The tests are performed with, and without, circular masks (as described above), and with systematically varied parameter settings (using grid search): initial Gaussian blur tested with values in the range $[0.4,2.4]$, with steps of 0.4; feature descriptor size tested with $\left\{1, 2, 4, 6, 8\right\}$;  steps per scale octave tested with $\left\{1, 2, 3, 4, 5\right\}$. The other available parameters are set to their default values, since we observed insensitivity to those parameters in our preliminary tests.

\subsubsection{Atlas-based Segmentation (LPBA40)}

In \cite{avants2011reproducible}, a protocol for evaluation of distance/similarity measures in the context of image registration was proposed. 
The protocol 
starts with affine registration, for which results are reported, and then proceeds to deformable registration. Since this study focuses on the development of an affine (linear) registration framework based on the proposed distance measure,  we compare with the reported affine-only performance; an improved affine registration is of great significance since a very high correlation between the performance of the affine registration and that of the subsequent deformable registration has been established.

We start from the two atlases created utilizing  the Advanced Neuroimaging Tools (ANTs) registration software suite and the open-source evaluation script provided in the reference study \cite{avants2011reproducible}. We utilize the atlas 
created using Mutual Information  
since that is the one found in \cite{avants2011reproducible} to be best performing and is used as the basis for the whole deformable registration study. Two-fold cross validation is utilized; the first atlas is registered to the last $20$ brain images and the second atlas is registered to the first $20$ brains, hence all registrations are done with brains that did not contribute to the creation of the atlas.

The multi-label segmentations defined by the atlas are transformed using the transformation parameters found during the registration and compared to the ground-truth segmentations for each brain. The Jaccard Index \cite{taha2015metrics} is calculated per region, as well as for the entire brain mask. 

For the proposed method based on $\dalphabidirssamdr$ we use $\alphalevelsconstant=7$, normalization percentiles $5\%$, $N=3000$, $0.05$ sampling fraction, and circular Hann windows as weight-masks. 

\begin{table}[t]
\centering
\caption{Registration of cilia: performance of the proposed method compared to reference
results, shown as the 'mean (std-dev)' of the registration error (in pixels) w.r.t. the considered sets of landmarks for the $19$ registrations. 'R' denotes rigid, 'A' denotes affine and 'D' dIenotes deformable registration. Bold marks the smallest error for each set of landmarks.}
\centering
\begin{tabular}{c|c||c|c|c}
\multicolumn{2}{c||}{Method} & \multicolumn{3}{c}{Registration Error}\\ \hline
Measure & Transform & Central Pair & Outer & All \\ \hline \hline
- & Identity & $4.32 \,(0.78)$ & $5.75 \,(3.49)$ & $5.61 \,(3.09)$ \\ \hline
\multirow{3}{*}{PCC \cite{suveer2017enhancement}} & R & $2.6 \,(1.5)$ & - & - \\
 & R+A & - & $3.27 \,(1.75)$ & - \\
 & R+A+D & $2.79 \,(1.84)$ & $2.30 \,(1.80)$ & $2.35 \,(1.82)$ \\ \hline
\multirow{2}{*}{$\dalphabidirsamdr$} & R & $2.65 \,(0.83)$ & $6.49 \,(2.64)$ & $6.10 \,(2.41)$ \\
 & R+A & $\mathbf{2.03 \,(1.04)}$ & $\mathbf{1.64 \,(0.36)}$ & $\mathbf{1.68 \,(0.29)}$ \\
\end{tabular}
\label{table:ciliareal}
\end{table}

\begin{table}[t!]
\caption{Results of  atlas-based brain segmentation. The table shows the Mean Jaccard Index for each of the brain regions for $\dalphabidirssamdr$ and mutual information with affine registration as reported in \cite{avants2011reproducible}. For $\dalphabidirssamdr$, mean and std. dev. are displayed; for the comparative results (MIAff), only mean was reported.}
\label{table:lpba40}
\centering
\resizebox{1.0 \columnwidth}{!} {
\begin{tabular}{|l|c|c|}
\hline
LPBA40 Label &\Bstrut\Tstrut $\dalphabidirssamdr$ & MIAff \cite{avants2011reproducible} \\ \hline
\Tstrut[2.3ex]All\_LPBA\_Data & \textbf{0.595} $\pm$ 0.0187  & 0.554 \\
Brain & \textbf{0.922} $\pm$ 0.0082  & 0.905 \\
\hline
21\_L\_superior\_frontal\_gyrus & 0.690 $\pm$ 0.0351  & \textbf{0.708} \\
22\_R\_superior\_frontal\_gyrus & 0.683 $\pm$ 0.0322  & \textbf{0.748} \\
23\_L\_middle\_frontal\_gyrus & \textbf{0.672} $\pm$ 0.0345  & 0.536 \\
24\_R\_middle\_frontal\_gyrus & \textbf{0.663} $\pm$ 0.0451  & 0.513 \\
25\_L\_inferior\_frontal\_gyrus & \textbf{0.590} $\pm$ 0.0467  & 0.569 \\
26\_R\_inferior\_frontal\_gyrus & \textbf{0.591} $\pm$ 0.0562  & 0.550 \\
27\_L\_precentral\_gyrus & \textbf{0.560} $\pm$ 0.0579  & 0.503 \\
28\_R\_precentral\_gyrus & \textbf{0.550} $\pm$ 0.0584  & 0.508 \\
29\_L\_middle\_orbitofrontal\_gyrus & \textbf{0.529} $\pm$ 0.0723  & 0.505 \\
30\_R\_middle\_orbitofrontal\_gyrus & \textbf{0.522} $\pm$ 0.0569  & 0.484 \\
31\_L\_lateral\_orbitofrontal\_gyrus & 0.434 $\pm$ 0.0828  & \textbf{0.551} \\
32\_R\_lateral\_orbitofrontal\_gyrus & 0.421 $\pm$ 0.0846  & \textbf{0.564} \\
33\_L\_gyrus\_rectus & \textbf{0.507} $\pm$ 0.0571  & 0.503 \\
34\_R\_gyrus\_rectus & \textbf{0.536} $\pm$ 0.0712  & 0.485 \\
41\_L\_postcentral\_gyrus & 0.479 $\pm$ 0.0739  & \textbf{0.490} \\
42\_R\_postcentral\_gyrus & \textbf{0.474} $\pm$ 0.0653  & 0.463 \\
43\_L\_superior\_parietal\_gyrus & \textbf{0.563} $\pm$ 0.0509  & 0.470 \\
44\_R\_superior\_parietal\_gyrus & \textbf{0.569} $\pm$ 0.0532  & 0.470 \\
45\_L\_supramarginal\_gyrus & 0.502 $\pm$ 0.0719  & \textbf{0.504} \\
46\_R\_supramarginal\_gyrus & \textbf{0.510} $\pm$ 0.0702  & 0.463 \\
47\_L\_angular\_gyrus & \textbf{0.509} $\pm$ 0.0782  & 0.506 \\
48\_R\_angular\_gyrus & \textbf{0.520} $\pm$ 0.0527  & 0.472 \\
49\_L\_precuneus & 0.525 $\pm$ 0.0613  & \textbf{0.546} \\
50\_R\_precuneus & \textbf{0.541} $\pm$ 0.0610  & 0.540 \\
61\_L\_superior\_occipital\_gyrus & \textbf{0.424} $\pm$ 0.0825  & 0.413 \\
62\_R\_superior\_occipital\_gyrus & \textbf{0.409} $\pm$ 0.0665  & 0.399 \\
63\_L\_middle\_occipital\_gyrus & \textbf{0.516} $\pm$ 0.0686  & 0.421 \\
64\_R\_middle\_occipital\_gyrus & \textbf{0.512} $\pm$ 0.0573  & 0.397 \\
65\_L\_inferior\_occipital\_gyrus & 0.448 $\pm$ 0.0967  & \textbf{0.484} \\
66\_R\_inferior\_occipital\_gyrus & 0.451 $\pm$ 0.0914  & \textbf{0.492} \\
67\_L\_cuneus & \textbf{0.442} $\pm$ 0.1087  & 0.372 \\
68\_R\_cuneus & \textbf{0.445} $\pm$ 0.0889  & 0.388 \\
81\_L\_superior\_temporal\_gyrus & \textbf{0.574} $\pm$ 0.0478  & 0.514 \\
82\_R\_superior\_temporal\_gyrus & \textbf{0.586} $\pm$ 0.0446  & 0.498 \\
83\_L\_middle\_temporal\_gyrus & \textbf{0.513} $\pm$ 0.0580  & 0.481 \\
84\_R\_middle\_temporal\_gyrus & \textbf{0.540} $\pm$ 0.0495  & 0.473 \\
85\_L\_inferior\_temporal\_gyrus & \textbf{0.509} $\pm$ 0.0601  & 0.462 \\
86\_R\_inferior\_temporal\_gyrus & \textbf{0.534} $\pm$ 0.0572  & 0.460 \\
87\_L\_parahippocampal\_gyrus & 0.546 $\pm$ 0.0743  & \textbf{0.556} \\
88\_R\_parahippocampal\_gyrus & 0.535 $\pm$ 0.0630  & \textbf{0.544} \\
89\_L\_lingual\_gyrus & \textbf{0.519} $\pm$ 0.0943  & 0.421 \\
90\_R\_lingual\_gyrus & \textbf{0.541} $\pm$ 0.0678  & 0.420 \\
91\_L\_fusiform\_gyrus & \textbf{0.542} $\pm$ 0.0801  & 0.488 \\
92\_R\_fusiform\_gyrus & \textbf{0.548} $\pm$ 0.0646  & 0.453 \\
101\_L\_insular\_gyrus & \textbf{0.625} $\pm$ 0.0504  & 0.378 \\
102\_R\_insular\_gyrus & \textbf{0.611} $\pm$ 0.0639  & 0.420 \\
121\_L\_cingulate\_gyrus & \textbf{0.553} $\pm$ 0.0504  & 0.491 \\
122\_R\_cingulate\_gyrus & \textbf{0.545} $\pm$ 0.0588  & 0.508 \\
161\_L\_caudate & \textbf{0.583} $\pm$ 0.1002  & 0.494 \\
162\_R\_caudate & \textbf{0.574} $\pm$ 0.0952  & 0.502 \\
163\_L\_putamen & \textbf{0.626} $\pm$ 0.0581  & 0.559 \\
164\_R\_putamen & \textbf{0.636} $\pm$ 0.0668  & 0.561 \\
165\_L\_hippocampus & 0.603 $\pm$ 0.0836  & \textbf{0.633} \\
166\_R\_hippocampus & 0.604 $\pm$ 0.0810  & \textbf{0.643} \\
181\_cerebullum & \textbf{0.813} $\pm$ 0.0416  & 0.659 \\
182\_brainstem & \textbf{0.778} $\pm$ 0.0447  & 0.660 \\ \hline
\end{tabular}
}
\end{table}

\subsection{Results of Real Applications} 
\subsubsection{Results of Registration of Cilia}

Performance of the proposed method, together with the best previously published results, are shown in Tab. \ref{table:ciliareal}. The table shows the mean and standard deviation of registration error (AME, in pixels) of the $19$ registrations, for the three considered sets of landmarks: the Central pair, the Outer rings, and All (1+9) ring pairs. 'R' denotes rigid; 'A' denotes affine; and 'D' denotes deformable registration.

The original study includes deformable registration as a final stage, after the rigid and affine steps. Here presented framework based on $\dalphabidirsamdr$ includes linear (rigid and affine),    but not deformable registration. However, as results included in Tab. \ref{table:ciliareal} confirm,  the proposed method outperforms the previous state-of-the-art, even if using only rigid and affine registrations.

We note that with only rigid registration we improve the alignment of the central pair while degrading the alignment of the outer rings. After the affine registration, the alignment of the central pair is improved further, plausibly due to the less constrained transformation model of affine compared to rigid, and we observe that the alignment of the outer rings and the total alignment are improved substantially.

The feature-based method is omitted from Tab. \ref{table:ciliareal} due to complete failure on all 19 image registration tasks, both with rigid and affine transformations; either too few matching points were detected, or the ones found resulted in large erroneous transformations. One such failed registration example is illustrated in Fig. \ref{fig:featurefail}.

\subsubsection{Results of Atlas-Based Segmentation of Brains}

Table \ref{table:lpba40} shows results of atlas-based brain segmentation. The Mean Jaccard Index is computed for each of the brain regions,  for $\dalphabidirssamdr$ and MI,  with affine registration as reported in \cite{avants2011reproducible}. For $\dalphabidirssamdr$, mean and std. dev. are displayed; for the comparative results (MIAff, \cite{avants2011reproducible}), only mean was reported.

We observe that for the whole brain mask, for the aggregated overlap, and for 43 out of the 56 distinct regions, the proposed measure outperforms the reported performance obtained with the MI  metric; MI  was the best performing measure out of the three evaluated in \cite{avants2011reproducible}.  

\section{Discussion}

Compared to the traditional similarity measures (SSD, PCC, MI), the proposed measure and associated registration method require substantial amounts of memory to store the auxiliary data-structures. A single 3D registration of two MR images of brains may require approximately 4GB of working memory with a reasonable set of parameters; contemporary machines for high-end data processing typically have a lot more memory than 4GB, but this requirement can affect how many registrations can be performed in parallel on a single machine.

\section{Conclusion}

In this study we have adapted a family of distance measures \cite{lindblad2014linear} to gradient descent based image registration, for 2D and 3D images. We have shown that such an extension is feasible and that the very good performance of the measures observed previously for object recognition and template matching, and their property of a large catchment basin for local optimization, also hold in the context of registration. This has been shown by evaluating the method in four main ways: (\romlcase{1}) on synthetic tests, (\romlcase{2}) execution time measurement, (\romlcase{3}) registration of TEM-images of cilia for multi-image super-resolution reconstruction, and (\romlcase{4}) atlas-based segmentation with annotated MR brain images. We observe that the proposed method provides outstanding performance for intensity-based affine registration in terms of robustness, accuracy and symmetry. It is also faster or similar in speed to the commonly used measures, which allows its practical applications.
The framework developed in this study operates on single-layer (e.g. gray-scale) images, but can be extended to multi-layer images such as color images, either by considering a linear sum of distances, or more sophisticated methods based on simultaneous presence or absence of membership in the multiple layers \cite{ofverstedt2017distance,sladoje2017distance}.
Future work includes extending the measures to non-linear (deformable), as well as multi-modal registration.

\section*{Acknowledgments}

The authors would like to thank Dr. Ida-Maria Sintorn for providing the cilia dataset and the landmarks used for the evaluation in Sec. \ref{sect:ciliaregistration}. This work is supported by VINNOVA, MedTech4Health grants 2016-02329 and 2017-02447, Swedish Research Council grants 2015-05878 and 2017-04385, and the Ministry of Education, Science, and Techn. Development of the Republic  of  Serbia  (proj.  ON174008  and  III44006).

\ifCLASSOPTIONcaptionsoff
  \newpage
\fi



\bibliographystyle{IEEEtran}
\bibliography{ImReg.bib}

\newpage

%
\begin{IEEEbiography}[{\includegraphics[width=1in,height=1.25in,clip,keepaspectratio]{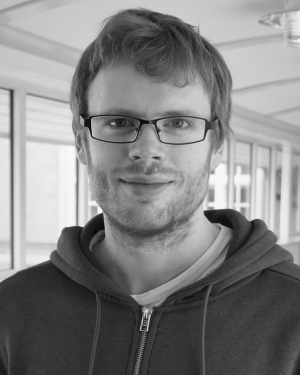}}]{Johan \"Ofverstedt}
received the B.Sc. and M.Sc. degrees in computer science in 2012 and 2017 respectively, and is currently pursuing a Ph.D. degree in computerized image processing at Centre for Image Analysis, Uppsala University, Sweden, since 2017. His research interests include distance measures, image registration, object recognition, and optimization.
\end{IEEEbiography}

\begin{IEEEbiography}[{\includegraphics[width=1in,height=1.25in,trim={0cm 0cm 0cm 0.3cm},clip,keepaspectratio]{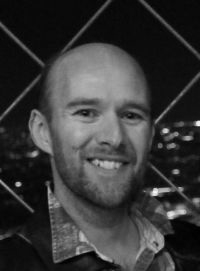}}]{Joakim Lindblad} received the M.Sc. in engineering physics and Ph.D. in computerized
image analysis from Uppsala University, Sweden, in 1997 and 2003, respectively. He is currently
Researcher at the Centre for Image Analysis, Uppsala University, Sweden; Senior Research Associate at
Mathematical Institute of the Serbian Academy of Sciences and Arts, Serbia; and Head of Research at Topgolf Sweden AB, Stockholm, Sweden. His research interests include development of general and robust methods for image processing and analysis.
\end{IEEEbiography}


\begin{IEEEbiography}[{\includegraphics[width=1in,height=1.25in,trim={5cm 0cm 6cm 0cm},clip,keepaspectratio]{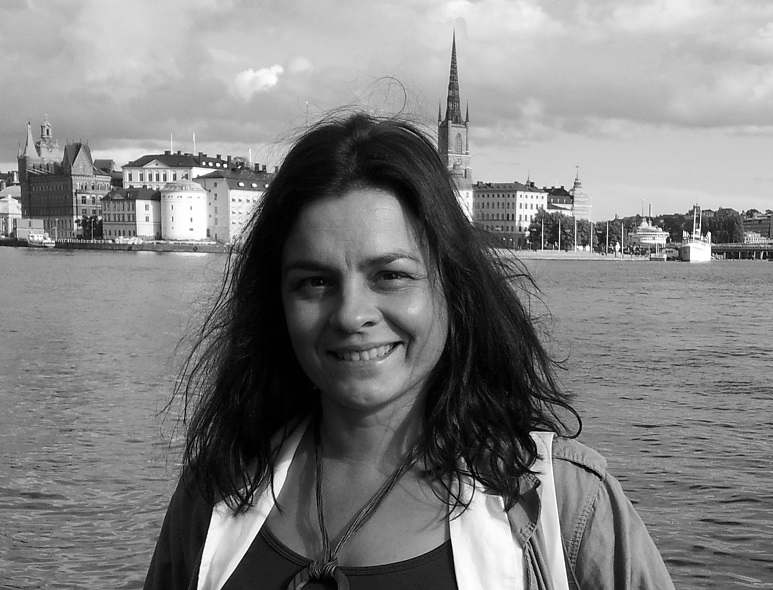}}]{Nata\v{s}a Sladoje}
received the B.Sc. and M.Sc. degrees in mathematics from the Faculty of Science,
University of Novi Sad, Serbia, in 1992 and 1998, respectively, and the Ph.D. degree in computerized
image analysis from the Centre for Image Analysis, Swedish University of Agricultural Sciences, Uppsala, Sweden, in 2005. She is a Senior Lecturer  at the Centre for Image Analysis, Uppsala University, Sweden. Her research interests include theoretical development of image analysis methods with applications in biomedicine and medicine. 
\end{IEEEbiography}




\end{document}